\documentclass[10pt,journal,compsoc]{IEEEtran}
\ifCLASSOPTIONcompsoc
  \usepackage[nocompress]{cite}
\else
  \usepackage{cite}
\fi
\ifCLASSINFOpdf
\else
\fi
\usepackage{amsmath,amsfonts,bm}

\newcommand{\figtop}{{\em (Top)}}
\newcommand{\figbottom}{{\em (Bottom)}}
\newcommand{\captiona}{{\em (a)}}
\newcommand{\captionb}{{\em (b)}}
\newcommand{\captionc}{{\em (c)}}

\def\Figref#1{Figure~\ref{#1}}

\def\Tabref#1{Table~\ref{#1}}

\def\Secref#1{Section~\ref{#1}}

\def\eqref#1{equation~\ref{#1}}

\def\1{\bm{1}}

\DeclareMathAlphabet{\mathsfit}{\encodingdefault}{\sfdefault}{m}{sl}
\SetMathAlphabet{\mathsfit}{bold}{\encodingdefault}{\sfdefault}{bx}{n}

\usepackage{hyperref}       
\usepackage{url}            
\usepackage{booktabs}       
\usepackage{amsfonts}       
\usepackage{nicefrac}       
\usepackage{microtype}      
\usepackage{xcolor}

\usepackage{subfig}
\usepackage{pifont}
\usepackage{graphicx}
\graphicspath{{figure/}}
\usepackage{enumitem}
\usepackage{wrapfig}
\usepackage{lipsum}
\usepackage{ulem}
\usepackage{mwe}

\newif\ifrevise
\revisefalse
\newcommand\changed[1]{\ifrevise \textcolor{blue}{#1} \else {#1}\fi}
\newcommand\removed[1]{\ifrevise \textcolor{red}{\sout{#1}} \else {} \fi}

\newif\ifdraft
\draftfalse

\newcommand{\ky}[1]{\ifdraft \textcolor{pink}{#1} \else {#1}\fi}
\newcommand{\KY}[1]{\ifdraft {\color{pink}{\bf #1}} \else {}\fi}

\newcommand{\ms}[1]{\ifdraft {\color{magenta}{#1}} \else {#1}\fi}

\newcommand{\comment}[1]{}

\newcommand{\Skdt}{Sparse Kendall-Tau}
\newcommand{\skdt}{sparse Kendall-Tau}
\newcommand{\pproxy}{$P_{proxy}$}
\newcommand{\fproxy}{$f_{proxy}$}
\newcommand{\pws}{$P_{ws}$}
\newcommand{\fws}{$f_{ws}$}

\newcommand\mypara[1]{\vspace{1mm}\noindent\textbf{#1}}

\begin{document}
%
\title{An Analysis of Super-Net Heuristics in Weight-Sharing NAS}
%
%
%
%

\author{Kaicheng~Yu,
        Ren\'{e}~Ranftl,
        and~Mathieu~Salzmann
\IEEEcompsocitemizethanks{		
\IEEEcompsocthanksitem Kaicheng~Yu is with the CVLab, EPFL and the Abacus.AI. \protect \\
E mail: kaicheng.yu.yt@gmail.com \protect \\
\IEEEcompsocthanksitem Mathieu~Salzmann is with the CVLab, School of Computer and Communication Sciences, EPFL. \protect\\
		E-mail: mathieu.salzmann@epfl.ch \protect\\
\IEEEcompsocthanksitem Rene~Ranftl is with the Intelligent Systems Lab, Intel. \protect \\
E-mail: rene.ranftl@intel.com
}
\thanks{Accepted to TPAMI, this version is for educational purpose only.}}

%
%

\markboth{Transaction of PAMI, August~2021}%
{Yu \MakeLowercase{\textit{et al.}}: Analyzing Super-Net Heuristics in WS-NAS}
%



\IEEEtitleabstractindextext{%
\begin{abstract}
Weight sharing promises to make neural architecture search (NAS) tractable even on commodity hardware.
Existing methods in this space rely on a diverse set of heuristics to design and train the shared-weight backbone network, a.k.a. the super-net. Since heuristics substantially vary across different methods and have not been carefully studied, it is unclear to which extent they impact super-net training and hence the weight-sharing NAS algorithms.
In this paper, we disentangle super-net training from the search algorithm, isolate 14 frequently-used training heuristics, and evaluate them over three benchmark search spaces. Our analysis uncovers that several commonly-used heuristics negatively impact the correlation between super-net and stand-alone performance, whereas simple, but often overlooked factors, such as proper hyper-parameter settings, are key to achieve strong performance. Equipped with this knowledge, we show that simple random search achieves competitive performance to complex state-of-the-art NAS algorithms when the super-net is properly trained.
\end{abstract}

}

\maketitle

\IEEEdisplaynontitleabstractindextext

%
\IEEEpeerreviewmaketitle

\IEEEraisesectionheading{\section{Introduction}\label{sec:intro}}

\begin{figure*}
    \centering
    \resizebox{0.65\textwidth}{!}{
    \includegraphics{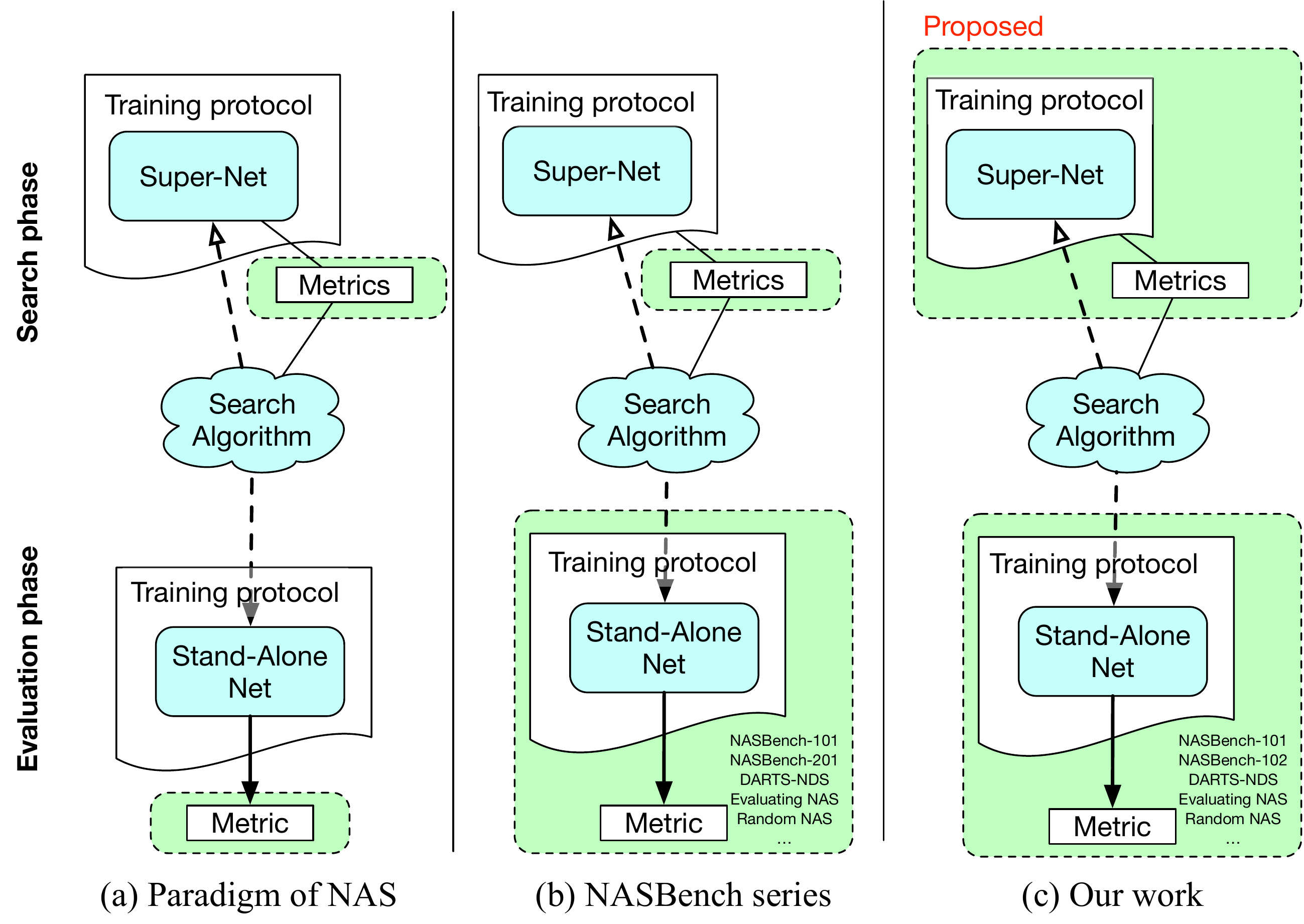}
    }
    \caption{
    \textbf{Weight Sharing NAS benchmarking.}
    Green blocks indicate which aspects of NAS are benchmarked in different works. A search algorithm usually consists of a search space that encompasses many architectures, and a policy to select the best one. 
    \removed{$P$ indicates a training protocol, and $f$ a mapping function from the search space to a neural network.}
    \textbf{(a)} 
    Early works fixed and compared the metrics on the proxy task, which doesn't allow for a holistic comparison between algorithms.
    \textbf{(b)} The NASBench benchmark series partially alleviates the problem by sharing the stand-alone training protocol and search space across algorithms. However, the design of the weight-sharing \changed{backbone network, a.k.a. super-net,} \removed{search space} and training protocol is still not controlled. 
    \textbf{(c)} We fill this gap by benchmarking existing techniques to construct and train the shared-weight backbone. We provide a controlled evaluation across three benchmark spaces. 
    }
    \label{fig:teaser}
\end{figure*}

\IEEEPARstart{N}{eural}
 architecture search~(NAS) has received growing attention in the past few years, yielding state-of-the-art performance on several machine learning tasks~\cite{liu2019autodeeplab,wu_fbnet:_2018,chen2019detnas,ryoo2020assemblenet}. One of the milestones that led to the popularity of NAS is weight sharing~\cite{Pham2018,Liu2018darts}, which, by allowing all possible network architectures to share the same parameters, has reduced the computational requirements from thousands of GPU hours to just a few. 
\Figref{fig:teaser} shows the two phases that are common to weight-sharing NAS (WS-NAS) algorithms: the search phase, including the design of the search space and the search algorithm; and the evaluation phase, which encompasses the final training protocol on the target task~\footnote{Target task refers to the tasks that neural architecture search aims to optimize on.}.

While most works focus on developing a good sampling algorithm~\cite{cai2018proxyless,xie2018snas} or improving existing ones~\cite{Zela2020Understanding,nayman2019xnas,li2019improving}, 
they tend to overlook or gloss over important factors related to the design and training of the shared-weight backbone network, i.e. the super-net. 
For example, the literature encompasses significant variations of learning hyper-parameter settings, batch normalization and dropout usage, capacities for the initial layers of the network, and depth of the super-net.
Furthermore, some of these heuristics are directly transferred from standalone network training to super-net training without carefully studying their impact in this drastically different scenario. For example, the fundamental assumption of batch normalization that the input data follows a slowly changing distribution whose statistics can be tracked 
during training is violated in WS-NAS, but nonetheless typically assumed to hold.

In this paper, we revisit and systematically evaluate commonly-used super-net design and training heuristics and uncover the strong influence of certain factors on the success of super-net training.
To this end, we leverage three benchmark search spaces, NASBench-101~\cite{ying2019bench},  NASBench-201~\cite{dong2020bench102}, and DARTS-NDS~\cite{radosavovic_network_2019}, for which the ground-truth stand-alone performance of a large number of architectures is available. We report the results of our experiments according to two sets of metrics: i) metrics that directly measure the quality of the super-net, such as the widely-adopted super-net accuracy~\footnote{\ky{The mean accuracy over a small set of randomly sampled architectures during super-net training.}} and a modified Kendall-Tau correlation between the searched architectures and their ground-truth performance, which we refer to as {\it \skdt{}}; ii)
proxy metrics
such as the ability to surpass random search and the stand-alone accuracy of the model found by the WS-NAS algorithm.


Via our extensive experiments (over 700 GPU days), we uncover that (i) the training behavior of a super-net  drastically differs from that of a standalone network, e.g., in terms of feature statistics and loss landscape, thus allowing us to define training settings, e.g., for batch-normalization~(BN) and learning rate, that are better suited for super-nets; (ii) while some neglected factors, such as the number of training epochs, have a strong impact on the final performance, others, believed to be important, such as path sampling, only have a marginal effect, and some commonly-used heuristics, such as the use of low-fidelity estimates, negatively impact it; 
(iii) the commonly-adopted super-net accuracy is unreliable to evaluate the super-net quality.

Altogether, our work is the first to systematically analyze the impact of the diverse factors of super-net design and training, and we uncover the factors that are crucial to design a super-net, as well as the non-important ones.
Aggregating these findings allows us to boost the performance of simple weight-sharing random search to the point where it reaches that of complex state-of-the-art NAS algorithms across all tested search spaces. Our code is available at \url{https://github.com/kcyu2014/nas-supernet}, and 
we will release our trained models so as to establish a solid baseline to facilitate further research.

\section{Preliminaries and Related Work}\label{sec:related}
We first introduce the necessary concepts that will be used throughout the paper. As shown in ~\Figref{fig:teaser}\captiona, weight-sharing NAS algorithms consist of three key components: a search algorithm that samples an architecture from the search space in the form of an encoding, a mapping function \fproxy{} that maps the encoding into its corresponding neural network, and a training protocol for a proxy task \pproxy{} for which the network is optimized.

To train the search algorithm, one needs to additionally define the mapping function \fws{} that generates the shared-weight network. Note that the mapping \fproxy{} frequently differs from \fws{}, since in practice the final model contains many more layers and parameters so as to yield competitive 
results on the proxy task. After fixing \fws{}, a training protocol \pws{} is required to learn the super-net. 
In practice, \pws{} often hides factors that are 
critical for the final performance of an approach, such as hyper-parameter settings or the use of data augmentation strategies to achieve state-of-the-art performance~\cite{Liu2018darts,chu_fairnas:_2019,Zela2020Understanding}. Again, \pws{} may differ from \pproxy{}, which is used to train the architecture that has been found by the search. For example, our experiments reveal that the learning rate and the total number of epochs frequently differ due to the different training behavior of the super-net and stand-alone architectures.

Many strategies have been proposed to implement the search algorithm, such as
reinforcement learning~\cite{Zoph2017,Zoph2018}, evolutionary algorithms~\cite{Real2017,Miikkulainen2019,so2019evolved,Liu2018b,Lu2018}, gradient-based optimization~\cite{Liu2018darts,Xu2020PC-DARTS:,li2019improving}, Bayesian optimization~\cite{kandasamy2018neural,jin2019auto,zhou2019bayesnas,wang2020neural}, and separate performance predictors~\cite{Liu2018b,Luo2018}.
Until very recently, the common trend to evaluate NAS consisted of reporting the searched architecture's performance on the proxy task~\cite{xie2018snas,real2018regularized,ryoo2020assemblenet}.
This, however, hardly provides real insights about the NAS algorithms themselves, because of the many 
components involved in them.
Many factors that differ from one algorithm to another can 
influence the performance. In practice, the literature even commonly compares NAS methods that employ different protocols to train the final model. 

\cite{li2019random} and \cite{yu2020evalnas} were the first to systematically compare different algorithms with the same 
settings for the proxy task and using several random initializations. Their surprising results revealed that many NAS algorithms produce architectures that do not significantly outperform a randomly-sampled architecture. 
\cite{Yang2020NAS} highlighted the importance of the training protocol \pproxy{}. They showed that optimizing the training protocol can improve the final architecture performance on the proxy task by three percent on CIFAR-10.
This non-trivial improvement can be achieved regardless of the chosen sampler, which provides clear evidence for the importance of unifying the protocol to build a solid foundation for comparing NAS algorithms. 

In parallel to this line of research, the recent series of ``NASBench" works~\cite{ying2019bench,Zela2020NAS-Bench-1Shot1,dong2020bench102} proposed to benchmark NAS approaches by providing a complete, tabular characterization of 
a search space. This was achieved by training every realizable stand-alone architecture using a fixed protocol \pproxy{}. Similarly, other works proposed to provide a partial characterization by sampling and training a sufficient number of architectures in a given search space using a fixed protocol~\cite{radosavovic_network_2019,Zela2020Understanding,wang2020neural}.


While recent advances for systematic evaluation are promising, no work has yet thoroughly studied the influence of the super-net training protocol \pws{} and the mapping function \fws{}. 
Previous works~\cite{Zela2020Understanding,li2019random} performed hyper-parameter tuning to evaluate their own algorithms, and focused only on a few parameters.
We fill this gap by benchmarking different choices of \pws{} and \fws{} and  by proposing novel variations to improve the super-net quality. 

\ky{Recent works have shown that \ms{sub-networks} of super-net training can surpass some human designed models without retraining~\cite{yu2020bignas,Cai2020Once} and that reinforcement learning can surpass the performance of random search~\cite{Bender2020tunas}. However, these findings are still only shown on MobileNet-like search spaces\ms{, where one only searches for the size of the} convolution kernels and the channel ratio for each layer. This is an effective approach to discover a compact network, but it does not change the fact that on more complex, cell-based search \ms{spaces the} super-net quality remains low. 
}

\section{Evaluation Methodology}

We first isolate 14 factors that need to be considered during the design and training of a super-net, and then introduce the metrics to evaluate the quality of the trained super-net. Note that these factors are agnostic to the search policy that is used after training the super-net.

\subsection{Disentangling the super-net design heuristics from the search algorithm}
\label{sec:method-factors}
\begin{figure}
    \centering
    \resizebox{\linewidth}{!}{
    \includegraphics{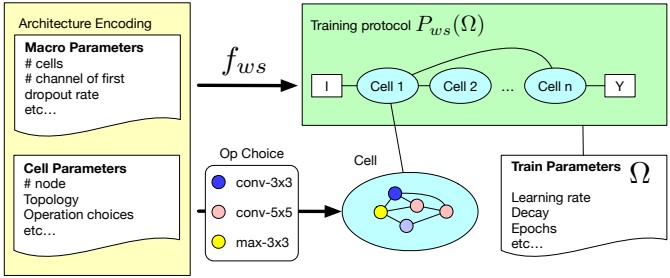}
    }\vspace{-0.1cm}
    \captionof{figure}{{\bf Constructing and training a super-net.}  \changed{Here, we formally introduce the components of super-net construction and training. 
     The super-net is a composition of all architectures within a search space. Each architecture is defined by a tabular architecture encoding. It consists of macro and cell parameters, which define the global structure and cell architecture, respectively.  We define the mapping function from a set of encodings into a super-net as $f_{ws}$ and the super-net training protocol given a set of  hyperparameters $\Omega$ as $P_{ws}$.
    We focus on cell-based search spaces that are widely adopted in the literature~\cite{Liu2018darts,Luo2018}. Note that some recent works remove the topology search within a cell to simplify the search space~\cite{cai2018proxyless} but can still be classified as cell-based. 
    }}
    \label{fig:construction} 
\end{figure}

Our goal is to evaluate the influence of the super-net mapping $f_{ws}$ and weight-sharing training protocol $P_{ws}$.
As shown in \Figref{fig:construction}, $f_{ws}$ translates an architecture encoding, which typically consists of a discrete number of choices or parameters, into a neural network. Based on a well-defined mapping, the super-net is a network in which every sub-path has a one-to-one mapping with an architecture encoding~\cite{Pham2018}. Recent works~\cite{Xu2020PC-DARTS:,li2019improving,ying2019bench} separate the encoding into \textit{cell parameters}, which define the basic building blocks of a network, and \textit{macro parameters}, which define how cells are assembled into a complete architecture.

\mypara{Weight-sharing mapping $f_{ws}$.}
To make the search space manageable, all cell and macro parameters are fixed during the search, except for the topology of the cell and its possible operations. However, the exact choices for each of these fixed factors differ between algorithms and search spaces. We 
report the
common factors in the left 
part of \Tabref{tab:factors}. They include various implementation choices, e.g., the use of convolutions with a dynamic number of channels~(Dynamic Channeling),  super-convolutional layers that support dynamic kernel sizes (OFA Kernel)~\cite{Cai2020Once}, weight-sharing batch-normalization~(WSBN) that tracks independent running statistics and affine parameters for different incoming edges~\cite{Luo2018}, and path and global dropout~\cite{Pham2018,Luo2018,Liu2018darts}. They also include the use of low-fidelity estimates~\cite{elsken2019neural} to reduce the 
complexity of super-net training,
e.g., by reducing the number of layers~\cite{Liu2018darts} and channels~\cite{Yang2020NAS,chen2019pdarts}, the portion of the training set used for super-net training~\cite{Liu2018darts}, 
or the batch size~\cite{Liu2018darts,Pham2018,Yang2020NAS}. 

\begin{table}[t]
    \centering
    \caption{\textbf{Summary of factors}}
    \resizebox{\linewidth}{!}{
\begin{tabular} {cc|cc}
\toprule
 \multicolumn{2}{c}{WS Mapping $f_{ws}$} & \multicolumn{2}{c}{WS Protocol $P_{ws}$}  \\
\cmidrule{1-2} \cmidrule{3-4}
 \textit{implementation} & \textit{low fidelity} &\textit{hyperparam.} & \textit{sampling} \\
\midrule
 Dynamic Channeling  & $\#$ layer & batch-norm & FairNAS  \\
OFA Conv & train portion & learning rate & Random-NAS \\
WSBN &  batch size & epochs & Random-A \\
Dropout & $ \# $ channels  & weight decay &  \\
Op on Node/Edge &&&\\
\bottomrule
\phantom{1}\\
\end{tabular}
} 
\label{tab:factors}
\end{table}

\mypara{Weight-sharing protocol $P_{ws}$}
Given a mapping $f_{ws}$, different training protocols $P_{ws}$ can be employed to train the super-net. Protocols can differ in the training hyper-parameters and the sampling strategies they rely on. We will evaluate the different hyper-parameter choices listed in the right part of \Tabref{tab:factors}. This includes the initial learning rate, the hyper-parameters of batch normalization, the total number of training epochs, and the amount of weight decay.

We randomly sample one path to train the super-net~\cite{guo_single_2019},
which is also known as single-path one-shot (SPOS) or Random-NAS~\cite{li2019random}. 
The reason for this choice is that Random-NAS is equivalent to the initial state of many search algorithms~\cite{Liu2018darts,Pham2018,Luo2018}, some of which even 
freeze the sampler training so as to use random sampling
to warm-up the super-net~\cite{Xu2020PC-DARTS:,dong2019searching}. Note that we also evaluated two variants of Random-NAS, but found their improvement to be only marginal. 

In our experiments, for the sake of reproducibility, we ensure that $P_{ws}$ and \pproxy{}, as well as $f_{ws}$ and \fproxy{}, are as close to each other as possible. For the hyper-parameters of $P_{ws}$, we cross-validate each factor following the order in \Tabref{tab:factors}, and after each validation, use the value that yields the best performance in \pproxy{}. 
For all other factors, we change one factor at a time.



\subsubsection{Search spaces} 
We employ three commonly-used \changed{cell-based} search spaces, for which a large number of stand-alone architectures have been trained and evaluated on CIFAR-10~\cite{Krizhevsky09cifar} to obtain their ground-truth performance. In particular, we use NASBench-101~\cite{ying2019bench}, which consists of $423,624$ architectures and is compatible with weight-sharing NAS~ \cite{yu2020evalnas,Zela2020NAS-Bench-1Shot1}; NASBench-201~\cite{dong2020bench102}, which contains more operations than NASBench-101 but fewer nodes; and DARTS-NDS~\cite{radosavovic_network_2019}  \ky{ that contains over $10^{13}$ architectures\ms{,} of which} a subset of 5000 models was sampled and trained in a stand-alone fashion. A summary of these search spaces and their properties is shown in \Tabref{tab:search-space}. The search spaces differ in the number of architectures that have known stand-alone accuracy~(\# Arch.), the number of possible operations~(\# Op.), how the  channels are handled in the convolution operations~(Channel), where dynamic means that the number of super-net channels might change based on the sampled architecture, and the type of optimum that is known for the search space~(Optimal). We further provide the maximum number of nodes ($n$), excluding the input and output nodes, in each cell, as well as a bound on the number of shared weights (Param.) and edge connections (Edges). Finally, the search spaces differ in how the nodes aggregate their inputs if they have multiple incoming edges (Merge). 

\changed{Recently, chain-like search spaces~\cite{cai2018proxyless,guo_single_2019,Xu2020PC-DARTS:,wu_fbnet:_2018} have been shown to be effective for computer vision tasks. However, to the best of our knowledge no benchmark space currently exists for such search spaces. We nonetheless construct a simplified chain-like space based on NASBench-101. See Appendix for more details.}
\begin{table}
    \centering
    \caption{\bf{Search Spaces}.}
    \vspace{-0.3cm}
    \resizebox{0.9\linewidth}{!}{
\begin{tabular} {l |ccc}
\toprule
& NASBench-101 & NASBench-201 & DARTS-NDS \\
\midrule
$\#$ Arch. & 423,624  & 15,625 & >$10^12$ \\
$\#$ Op. & 3 & 5 & 8 \\
Channel & Dynamic & Fix & Fix \\
Optimal & Global & Global & Sample \\
Nodes=($n$) & 5 & 4 & 4 \\
Param. & $O(n)$ & $O(n)$ - $O(n^2)$ & $O(n)$ - $O(n^2)$ \\
Edges & $O(n^2)$ & $O(n^2)$ & $O(n)$ \\
Merge & Concat. & Sum & Sum \\
\bottomrule
\end{tabular}
}\vspace{-0.4cm}
\label{tab:search-space}
\end{table}

\subsection{\Skdt{} - A novel super-net evaluation metric}
\begin{figure}
    \centering
    \vspace{0.5cm}
    \begin{minipage}{.7\linewidth}
    \resizebox{\linewidth}{!}{
    \includegraphics{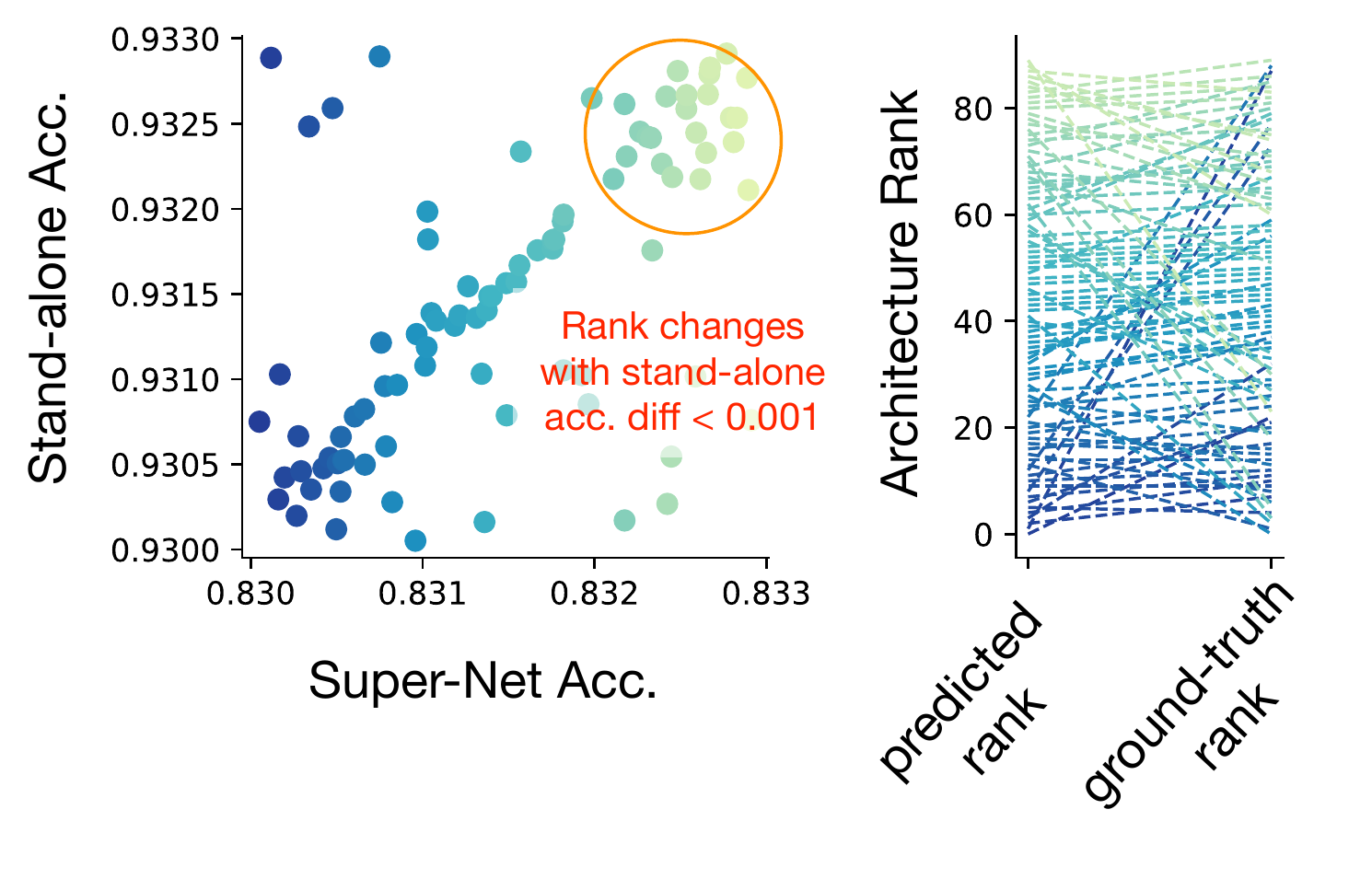}
    }
    \vspace{-0.5cm}
    \end{minipage}\hfill
    \begin{minipage}{0.3\linewidth}
    \vspace{-1.0cm}
    \resizebox{\linewidth}{!}{
    		\begin{tabular}{l|c}
			\toprule
			  &  Kendall\\
			  & Tau \\
			  \midrule
			  original  & 0.6444 \\
			  sparse &  0.8140  \\
			\bottomrule
    \end{tabular}
		}
    \end{minipage}
    \vspace{-0.35cm}
    \caption{\textbf{Kendall-Tau vs \Skdt{}.} 
    Kendall-Tau is not robust when many architectures have similar performance.
    Minor performance differences can lead to large perturbations in the ranking. 
    Our \skdt{} alleviates this by dismissing minor differences in performance.
    } 
    \label{fig:sparse-kdt}
    \vspace{-0.4cm}
\end{figure}


We define a novel super-net metric, which we name \textit{\skdt{}}. It is inspired by the Kendall-Tau metric used by \cite{yu2020evalnas} to measure the discrepancy between the ordering of stand-alone architectures and the ordering that is implied by the trained super-net. An ideal super-net should yield the same ordering of architectures as the stand-alone one and thus would lead to a high Kendall-Tau. However, Kendall-Tau is not robust to negligible performance differences between architectures (c.f. \Figref{fig:sparse-kdt}). To robustify this metric, we share the rank between two architectures if their stand-alone 
accuracies differ by less than a threshold ($0.1\%$ here). Since the resulting ranks are sparse, we call this metric \textit{\skdt{}} (s-KdT). See Appendix for implementation details.

\mypara{Sparse Kendall-Tau threshold}.
This value should be chosen according to what is considered a significant improvement for a given task. For CIFAR-10, where accuracy is larger than 90\%, we consider a 0.1\% performance gap to be sufficient. For tasks with smaller state-of-the-art performance, larger values might be better suited.

\mypara{Number of architectures}.
In practice, we observed that the \skdt{} metric became stable and reliable when using at least $n=150$ architectures. We used $n=200$ in our experiments to guarantee stability and fairness of the comparison of the different factors.

\mypara{Limitation of Sparse Kendall-Tau}
We nonetheless acknowledge that our \skdt{} has some limitations. For example, a failure case of using \skdt{} for super-net evaluation may occur when the top 10\% architectures are perfectly ordered, while the bottom 90\% architectures are purely randomly distributed. In this case, the Kendall Tau will be close to 0. However, the search algorithm will always return the best model, as desired. 

Nevertheless, while this corner case would indeed be problematic for the standard Kendall Tau, it can be circumvented by tuning the threshold of our sKdT. A large threshold value will lead to a small number of groups, whose ranking might be more meaningful. For instance in some randomly-picked NASBench-101 search processes, setting the threshold to 0.1\% merges the top 3000 models into 9 ranks, but still yields an sKdT of only 0.2. Increasing the threshold to 10\% clusters the 423K models into 3 ranks, but still yields an sKdT of only 0.3. This indicates the stability of our metric.

\begin{figure}[t]
    \centering
    \vspace{-0.2cm}
    \resizebox{\linewidth}{!}{
    \includegraphics{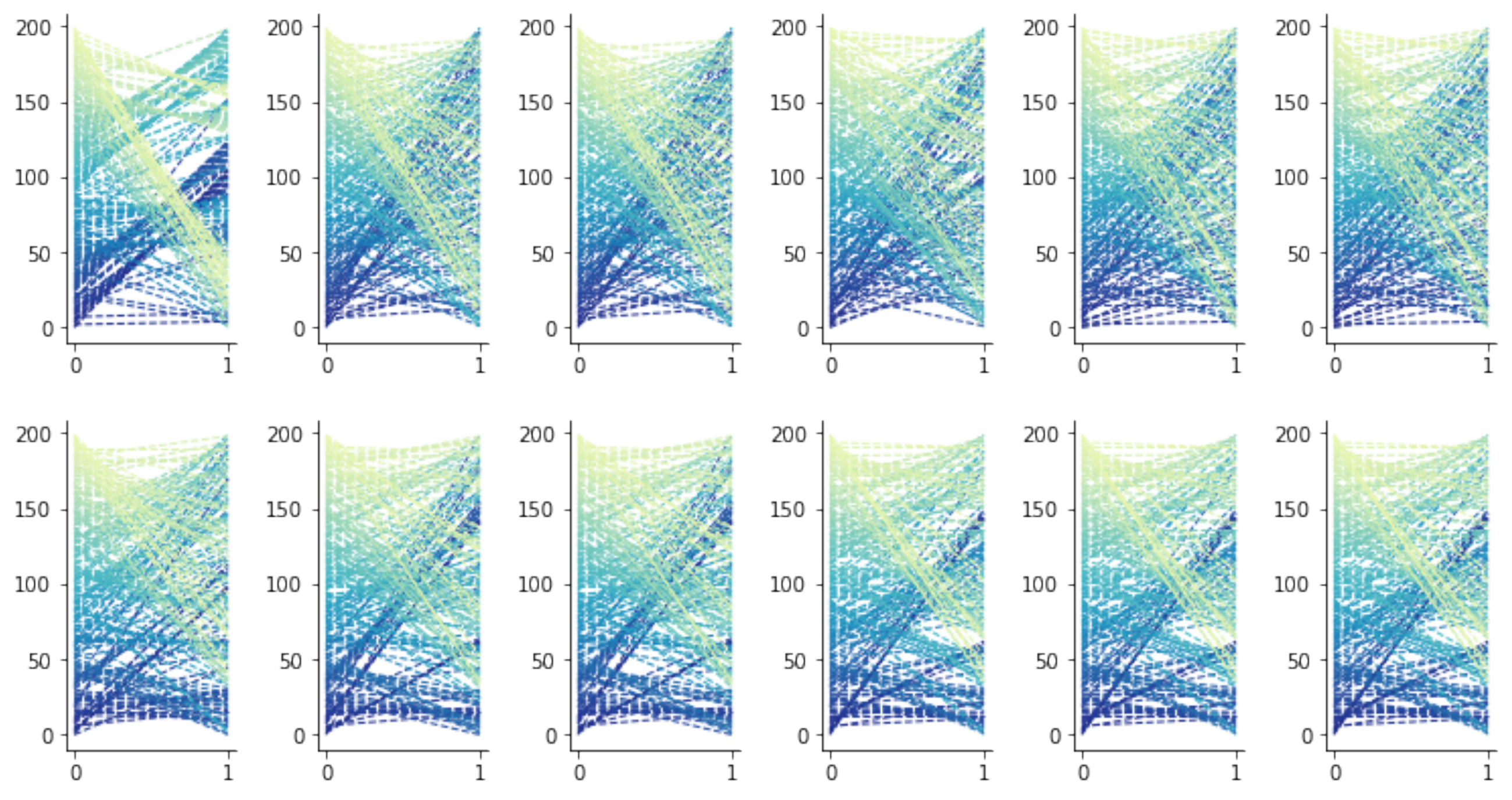}
    }
    \vspace{-0.5cm}
    \caption{
    \textbf{Ranking disorder examples.} We randomly select 12 runs from our experiments. For each sub-plot, 0 indicates the architecture ground-truth rank, and 1 indicates the ranking according to their super-net accuracy. 
    \changed{These plots evidence} that the ranking disorder does not follow a particular pattern.
    }
    \label{fig:rank-disorder-example}
\end{figure}

In \Figref{fig:rank-disorder-example}, we randomly picked 12 settings and show the corresponding bipartite graphs relating the super-net and ground-truth rankings to investigate where disorder occurs. In practice, the corner case discussed above virtually never occurs; the ranking disorder is typically spread uniformly across the architectures.

\subsubsection{Other metrics} 
\label{sec:other-metrics}
Although, \skdt{} captures the super-net quality well, it may fail in extreme cases, such as when the top-performing architectures are ranked perfectly while poor ones are ordered randomly. To account for such rare situations and ensure the soundness of our analysis, we also report additional metrics.
We define two groups of metrics to holistically evaluate different aspects of a trained super-net.

The first group of metrics directly evaluates the quality of the super-net, including \skdt{} and the widely-adopted super-net accuracy.
For the \emph{super-net accuracy}, we report the average accuracy of
200 architectures on the validation set of the dataset of interest. We will refer to this metric simply as \textit{accuracy}. It is frequently used \cite{guo_single_2019,chu_fairnas:_2019} to assess the quality of the trained super-net, but we will show later that it is in fact a poor predictor of the final stand-alone performance.
The metrics in the second group evaluate the search performance of a trained super-net. 
The first metric is the \emph{probability to surpass random search}:
Given the ground-truth rank $r$ of the best architecture found after $n$ runs and the maximum rank $r_{max}$, equal to the total number of architectures, the probability that the best architecture found is better than a randomly searched one 
is given by $p = 1 - (1 - (r / r_{max}))^n$.

Finally, where appropriate, we report the \emph{stand-alone accuracy of the model,} \changed{a.k.a. final performance}, that was found by the complete WS-NAS algorithm. Concretely, we randomly sample 200 architectures, select the 3 best models based on the super-net accuracy and query the ground-truth performance. We then take the mean of these architectures as stand-alone accuracy. Note that the same architectures are used to compute the \skdt{}.

\section{Analysis}
\label{sec:analysis}
We provide an analysis on the impact of the factors that are shown in \Tabref{tab:factors} across three different search spaces. In addition, we report the complete numerical results of all metrics in Section~\ref{apdx:allfactor}.

\mypara{Training Details.}
We use PyTorch~\cite{pytorch} for our experiments. Since NASBench-101 was constructed in TensorFlow we implement a mapper that translates TensorFlow parameters into our PyTorch model. We exploit two large-scale experiment management tools, SLURM~\cite{slurm} and Kubernetes~\cite{k8s}, to deploy our experiments.
We use various GPUs throughout our project, including NVIDIA Tesla V100, RTX 2080 Ti, GTX 1080 Ti and Quadro 6000 with CUDA 10.1.
Depending on the number of training epochs, parameter sizes and batch-size, most of the super-net training finishes within 12 to 24 hours, with the exception of FairNAS, whose training time is longer, as discussed earlier. 
We split the data into training/validation using a 90/10 ratio for all experiments, except those involving validation on the training portion. Please consult our submitted code for more details.

\mypara{Reproducing the Ground Truth from Tensorflow.}
\begin{figure}
    \centering
    \resizebox{\linewidth}{!}{
    \includegraphics{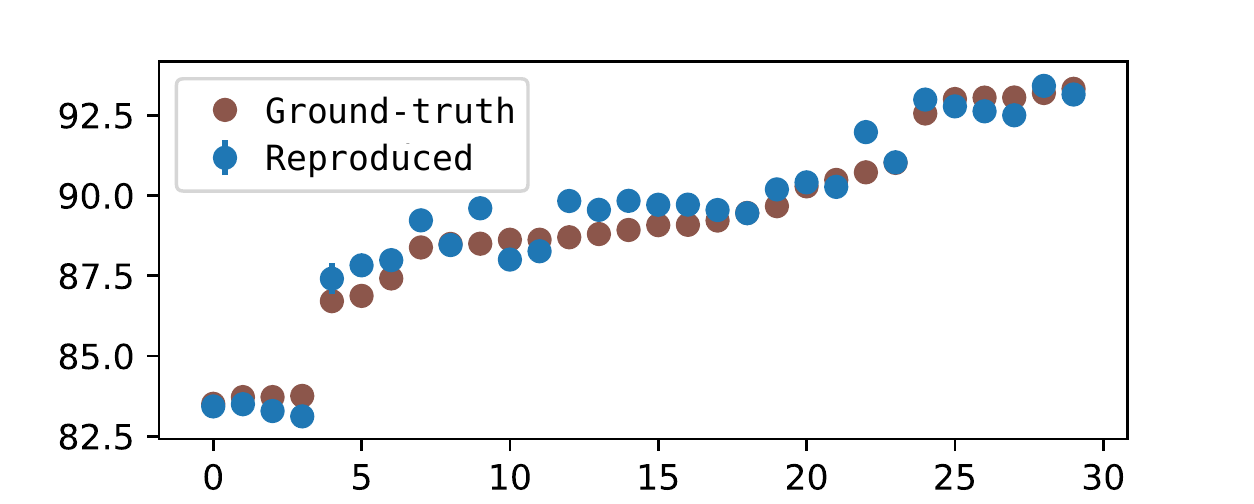}
    }
    \vspace{-0.5cm}
    \caption{Reproducing NASBench-101.}
    \label{fig:reproduce}
    \vspace{-0.4cm}
\end{figure}
As the ground-truth performance used by NASBench-101 are obtained on Tensorflow with TPU computation structure. We firstly reproduce these results in Pytorch with our implementation to make sure the re-implementation is trustworthy.
We uniformly random sampled 10 architectures, and repeat 3 times. It results 30 architectures and covers the spectrum of performance from 82\% to 93\%. 
We adopted the optimizer and hyper-parameter setting according to the code release of NASBench-101, repeated with 3 random initializations and take the mean performance. \changed{Note that we copied the initialization from the released Tensorflow model into PyTorch format to minimize frameworks discrepancies.}
We plot the performance comparison in \Figref{fig:reproduce}. The Kendall Tau metric is 0.81, and should be considered as the upper-bound of super-net training. It clearly indicates that even the reproducing results are not perfectly aligned with the Tensorflow original, it cannot explain why the significant drop to 0.2 after using weight sharing~\cite{yu2020evalnas}.

\begin{figure}
    \centering
    \resizebox{\linewidth}{!}{
    \hspace{-1.cm}
    \includegraphics{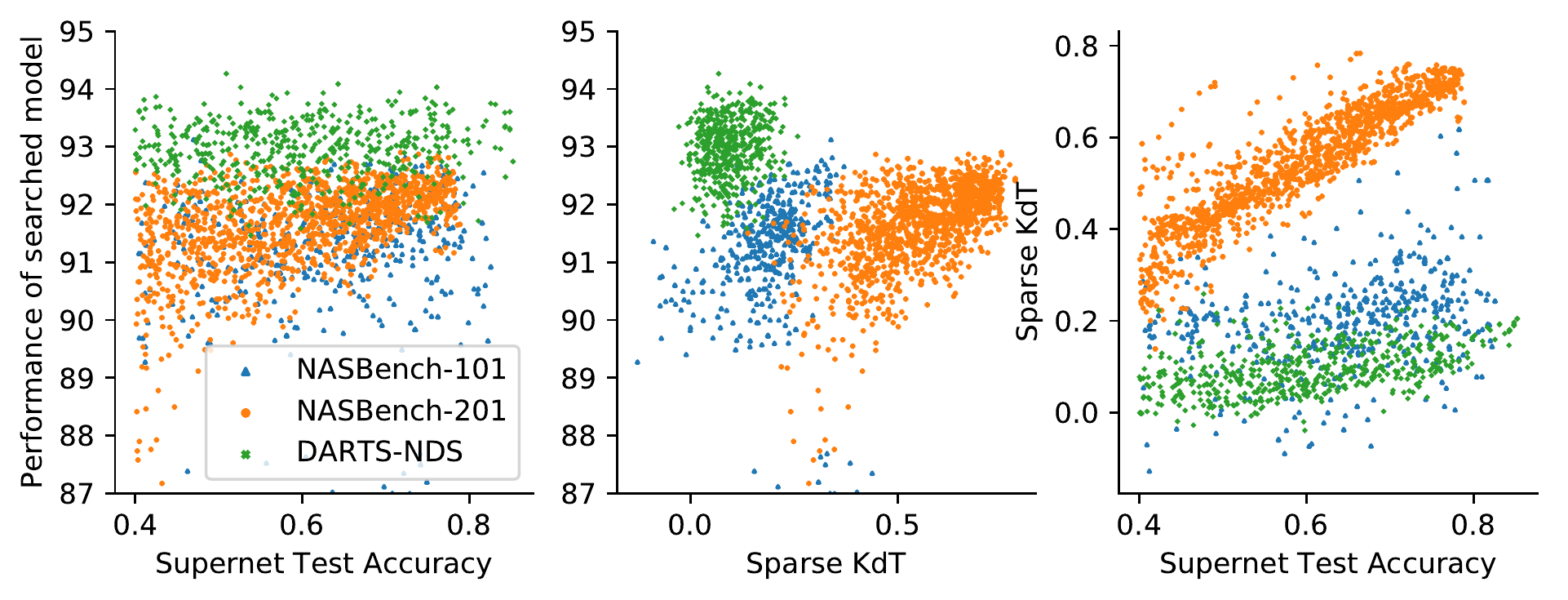}
    }
    \resizebox{0.8\linewidth}{!}{
    \begin{tabular}{l|ccc}
    Spearman Corr.     &  Acc. - Perf. & S-KdT - Perf. & Acc. - S-KdT \\
    \midrule
    NASBench-101 & 0.09  & 0.45 & 0.23 \\
    NASBench-201 & 0.52  & 0.55 & 0.94 \\
    DARTS-NDS    & 0.07  & 0.19 & 0.47 \\
    \end{tabular}}
    \vspace{-0.2cm}
    \caption{\textbf{Super-net evaluation}. We collect all experiments across 3 benchmark spaces. \textbf{(Top)} Pairwise plots of super-net accuracy, final performance, and the \skdt{}. Each point corresponds to  
    statistics computed over a trained super-net. \changed{(See Section~\ref{sec:other-metrics} for more details about metrics computation.)} Note that for super-net accuracy, we filter the super-net accuracy below 40\% to avoid statistics of ill-trained super-net.
    \textbf{(Bottom)} Spearman correlation coefficients between the metrics.
    }
    \label{fig:supernet}
\end{figure}

\subsection{Evaluation of a super-net}
\label{subsec:supernet-eval}

The standalone performance of the architecture that is found by a NAS algorithm is clearly the most important metric to judge its merits. 
However, in practice, one cannot access this metric---we wouldn't need NAS if standalone performance was easy to query (the cost of computing stand-alone performance is discussed in Section~\ref{apdx:saa-vs-skt}). 

Furthermore, stand-alone performance inevitably depends the sampling policy, and does not directly evaluate the quality of the super-net (see Section~\ref{apdx:saa-vs-skt}).
Consequently, it is important to 
rely on metrics that are well correlated with the final performance but can be queried efficiently. 
To this end, we collect all our experiments and plot the pairwise correlation between final performance, \skdt{}, and super-net accuracy. As shown in \Figref{fig:supernet}, the super-net accuracy has a low correlation with the final performance 
on NASBench-101 and DARTS-NDS. Only on NASBench-201 does it reach a correlation of 0.52.
The \skdt{} yields a consistently higher correlation with the final performance. This is evidence that one should not focus too strongly on improving the super-net accuracy.
While this metric remains computationally heavy, it serves as a middle ground that is feasible to evaluate in real-world applications. 

In the following experiments,
we thus mainly rely on \skdt{}, and use final search performance as a reference only.

\begin{figure*}[t]
    \centering
    \resizebox{0.75\linewidth}{!}{
    \includegraphics{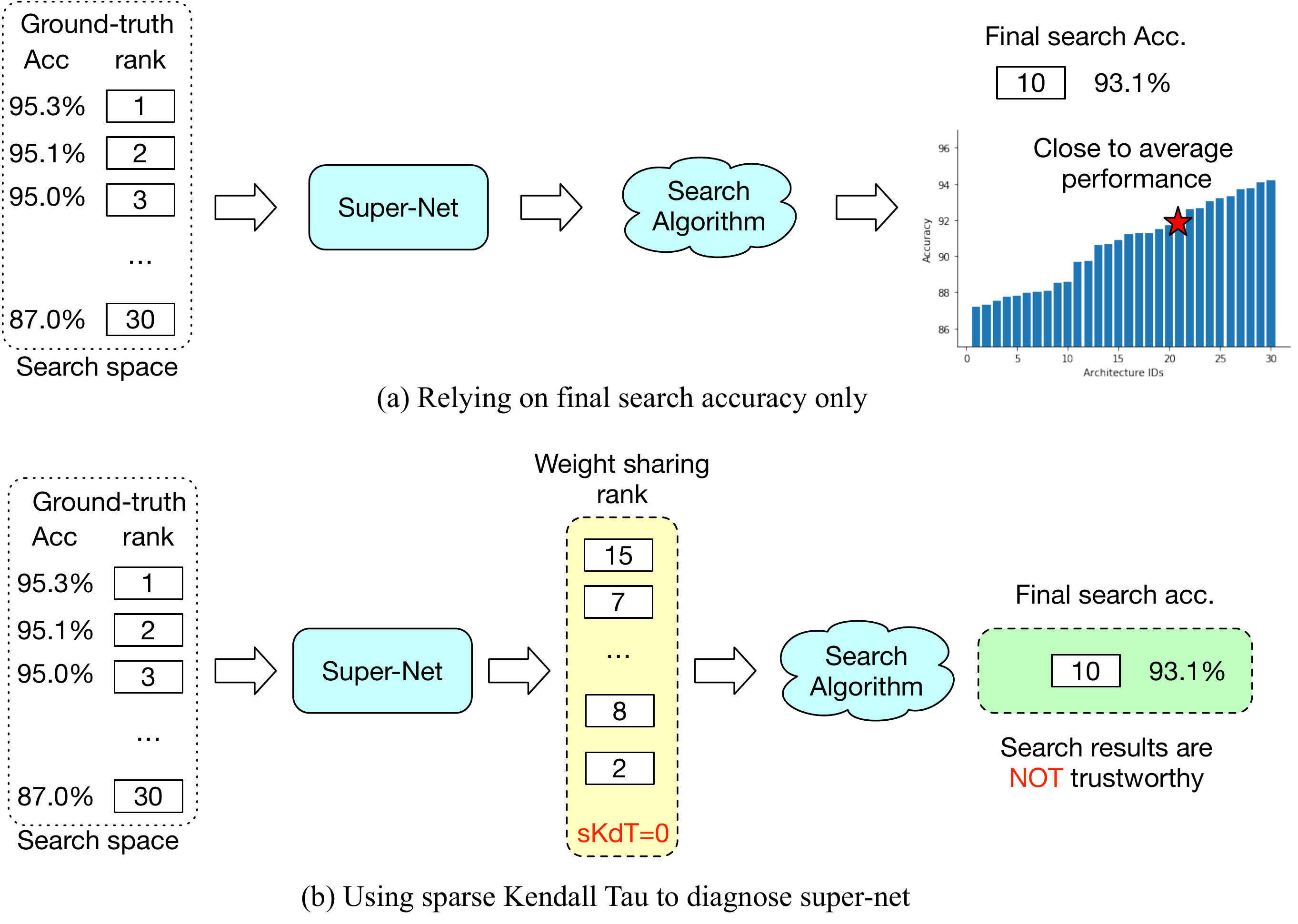}
    }
    \vspace{-0.3cm}
    \caption{\textbf{Comparing \skdt{} and final search accuracy.} Here, we provide a toy example to illustrate why one cannot rely on the final search accuracy to evaluate the quality of the super-net. Let us consider a search space with only 30 architectures, whose accuracy ranges from 95.3\% to 87\% on the CIFAR-10 dataset, and we run a search algorithm on top. \textbf{(a)} describes a common scenario: we run the search for multiple times, yielding a best architecture with 93.1\% accuracy. While this may seem good, it does not give any information about the quality of the search or the super-net. If we had full knowledge about the performance of every architecture in this space, we would see that this architecture is close to the average performance and hence no better than random. 
    In \textbf{(b)}, the \skdt{} allows us to diagnose this pathological case. A small \skdt{} implies that there is a problem with super-net training.
    }
    \label{fig:skdt-fsa}
\end{figure*}

\subsubsection{Kendall Tau v.s. Spearman ranking correlation}
\begin{table}[t]
\captionof{table}{Comparison of Kendall Tau~(KdT) and Spearman ranking (SpR) with their sparse variants.}
\centering
\resizebox{0.8\linewidth}{!}{
 \begin{tabular}{l|cccc} 
        \toprule
        Corr. of Perf.    &  KdT. & S-KdT & SpR & S-SpR\\
        \midrule
        NASBench-101 & 0.29  & 0.45 & 0.23 & 0.41 \\
        NASBench-201 & 0.42  & 0.55 & 0.38 & 0.57 \\
        DARTS-NDS    & 0.08  & 0.19 & 0.09 & 0.20 \\
        \bottomrule
        \end{tabular}
        }
        \vspace{-0.2cm}
        \label{tab:metrics}
        \vspace{-0.2cm}
\end{table}
\label{subsubsection:kdt-spr}

Kendall-tau is not the only metric to evaluate the ranking correlation. Spearman ranking correlation is also widely adopted in this field~\cite{guo_single_2019,dong2020bench102}. Note that our idea of sparsity also applies to SpR. In \Tabref{tab:metrics}, we compare the performance of Kendall Tau(KdT), Spearman ranking correlation (SpR) and their sparse variants, in the same setting as \Figref{fig:supernet}. Note that SpR and KdT performs similarly but that their sparse variants effectively improve the correlation on all search spaces.

\subsubsection{Stand-alone Accuracy v.s. Sparse Kendall-Tau}
\label{apdx:saa-vs-skt}
A common misconception is that the super-net quality is well reflected by stand-alone accuracy of the final selected architecture. Neither \skdt{} (sKdT) nor stand-alone accuracy are perfect. Both are tools to measure different aspects of a super-net.

Let us consider a completely new search space in which we have no prior knowledge about performance. As depicted by \Figref{fig:skdt-fsa}, if we only rely on the stand-alone accuracy, the following situation might happen:
Due to the lack of knowledge, the ranking of the super-net is purely random, and the search space accuracy is uniformly distributed. When trying different settings, there will be 1 configuration that `outperforms’ the others in terms of stand-alone accuracy. However, this configuration will be selected by pure chance. By only measuring stand-alone accuracy, it is technically impossible to realize that the ranking is random. 
By contrast, if one measures the sKdT (which is close to 0 in this example), an ill-conditioned super-net can easily be identified. In other words, purely relying on stand-alone accuracy could lead to pathological outcomes that can be avoided using \skdt{}.

Additionally, stand-alone accuracy is related to both the super-net and the search algorithm. \skdt{} allows us to judge super-net accuracy independently from the search algorithm. As an example, consider the use of a reinforcement learning algorithm, instead of random sampling, on top of the super-net. When observing a poor stand-alone accuracy, one cannot conclude if the problem is due to a poor super-net or to a poor performance of the RL algorithm. Prior to our work, people relied on the super-net accuracy to analyze the super-net quality. This is not a reliable metric, as shown in \Figref{fig:supernet}.  We believe that \skdt{} is a better alternative. 

\mypara{Computational Cost.} Computing the final accuracy is more expensive than training the super-net. Despite the low-fidelity heuristics reducing the weight-sharing costs, training a stand-alone network to convergence has higher cost, e.g., DARTS searches for 50 epochs but trains from scratch for 600 epochs~\cite{Liu2018darts}. Furthermore, debugging and hyper-parameter tuning typically require training thousands of stand-alone models. Note that, as one typically evaluates a random subset of architectures to understand the design space~\cite{radosavovic_network_2019}, \skdt{} can be computed without additional costs. In any event, the budget for \skdt{} is bounded with $n$.

\subsection{Weight-sharing Protocol \pws{} -- Hyperparameters}
\begin{figure}[t]
    \centering
    \resizebox{0.85\linewidth}{!}{
    \includegraphics{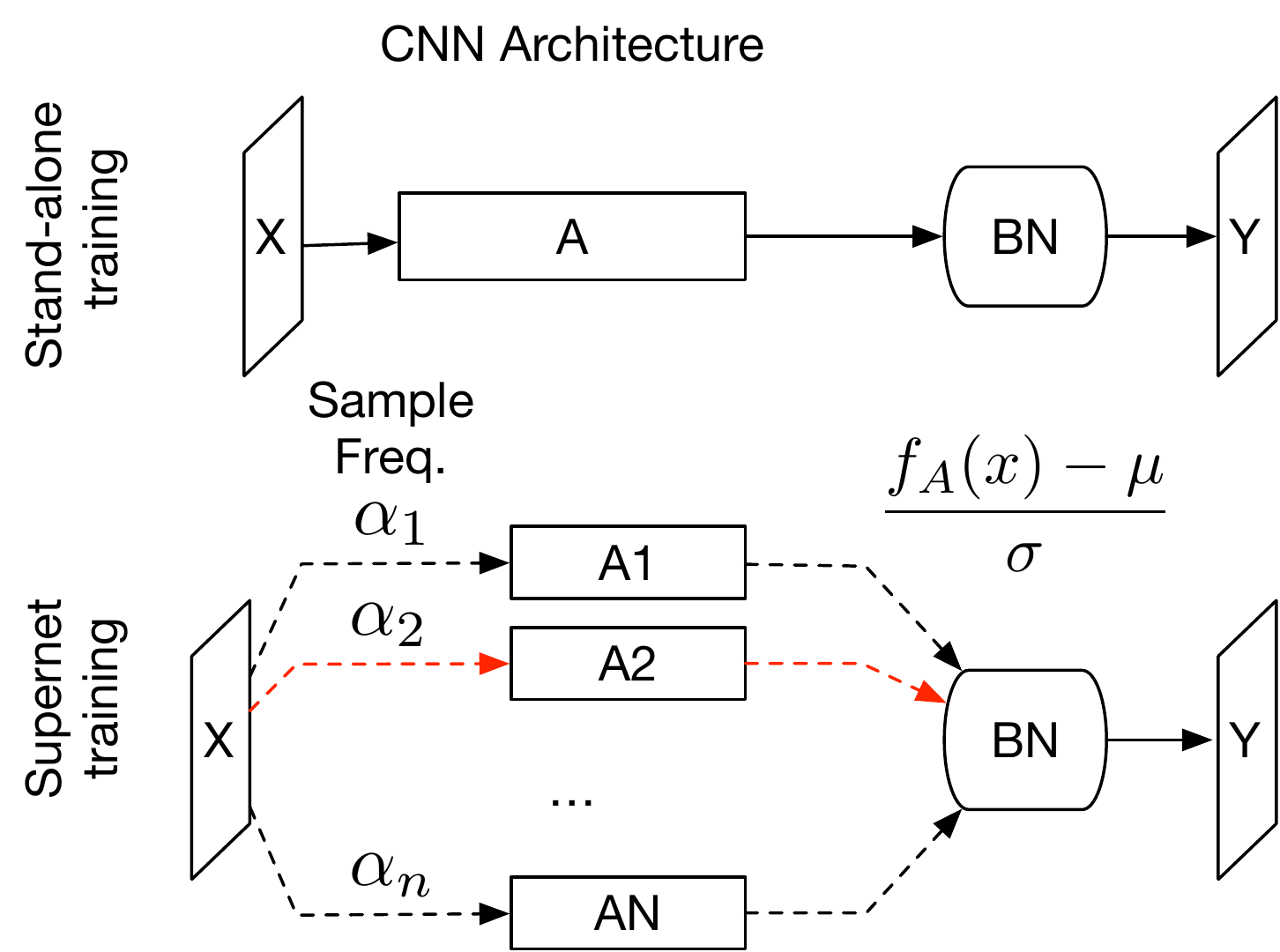}
    } 
    \vspace{-0.1cm}
    \caption{Batch normalization in standalone and super-net training.}
    \label{fig:bn-toy}
    \vspace{0.1cm}
    \resizebox{\linewidth}{!}{
    \includegraphics{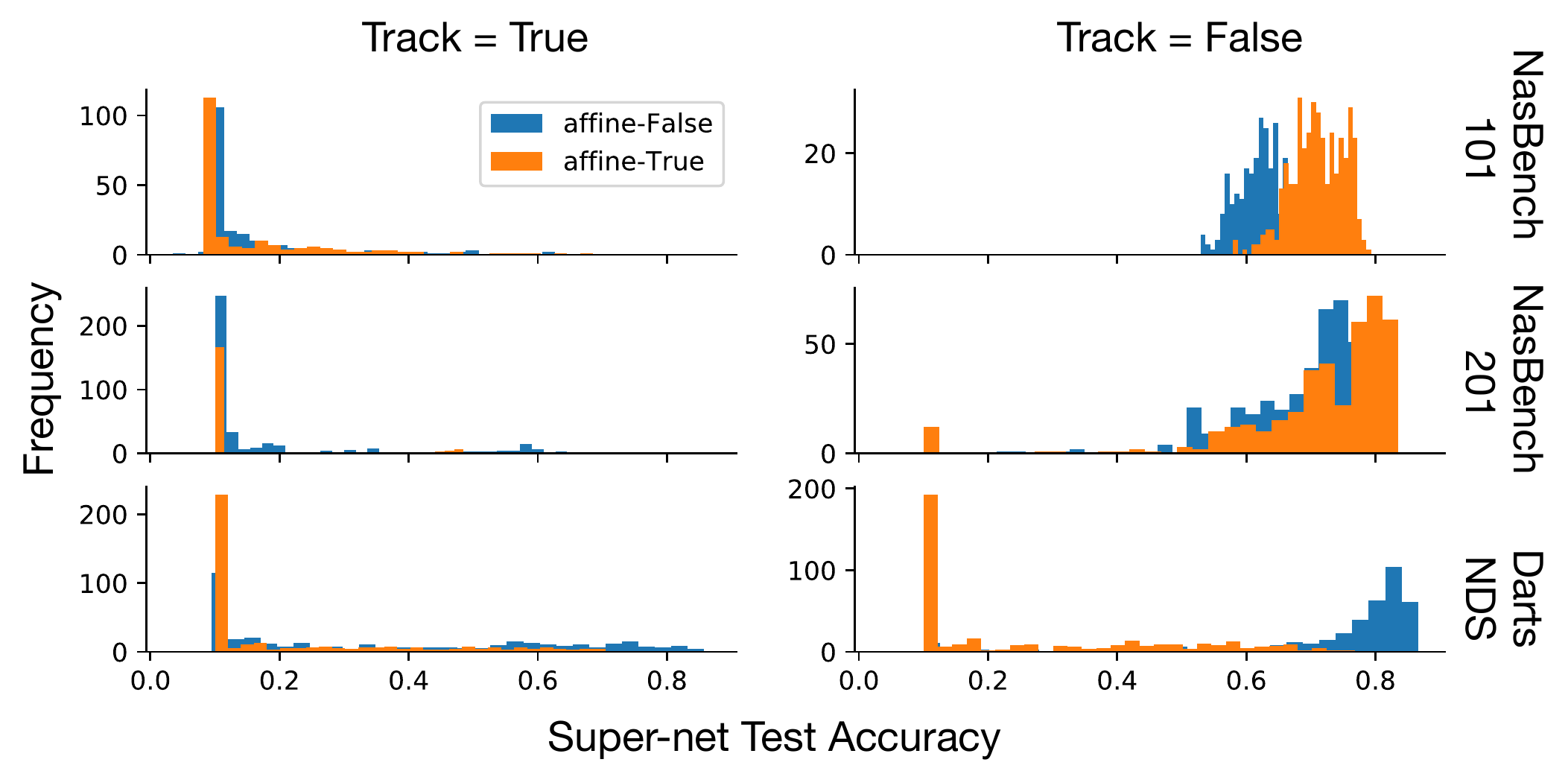}
    }
    \vspace{-0.6cm}
    \caption{\textbf{Validation of BN.}
    We plot histograms of the super-net accuracy for different hyper-parameter settings. Tracking statistics \textbf{(left)} leads to many architectures with random performance. Without tracking \textbf{(right)}, learning the affine parameters (\textit{affine-true}) increases accuracy on NASBench-101 and NASBench-201, but strongly decreases it for DARTS-NDS.
    }
    \label{fig:bn}
    \vspace{-0.5cm}
\end{figure}


\subsubsection{Batch normalization in the super-net}
\label{subsec:batchnorm}

%

Batch normalization (BN) is commonly used in standalone networks to allow for faster and more stable training. It is thus also employed in most CNN search spaces. However, BN behaves differently in the context of WS-NAS, and special care has to be taken when using it. 
In a standalone network (c.f. \Figref{fig:bn-toy}~\figtop), a BN layer during training computes the batch statistics $\mu_B$ and $\sigma_B$, normalizes the activations $f_{A}(x)$  as $(f_A(x)-\mu_B) / \sigma_B$, and finally updates the population statistics using a moving average. For instance, the mean statistics is updated as $\hat{\mu} \leftarrow \gamma\hat{\mu} + (1-\gamma) \mu_B$. At test time, the stored population statistics are used to normalize the feature map. 
In the standalone setting, both batch and population statistics are unbiased estimators of the population distribution $\mathcal{N}(\mu, \sigma)$. 



By contrast, when training a super-net (\Figref{fig:bn-toy}~\figbottom) the population statistics that are computed based on the running average are not unbiased estimators of the population distribution, because the effective architecture before the BN layer varies in each epoch. More formally, let $f_{A_i}$ denote the $i$-th architecture. During training, the batch statistics are computed as $\mu_B^i = \sum_j f_{A_i} (x_j) / m$, and the output feature follows the distribution $\mathcal{N}(\mu_B^i, \sigma_B^i)$, where the superscript $i$ indicates that the current batch statistics depends on $A_i$ only.
The population mean statistics is then updated as $\hat{\mu} \leftarrow \gamma \hat{\mu} + (1-\gamma)\mu_B^i$. However, during training, different architecture from the super-net are sampled. Therefore, the population mean statistics essentially becomes a weighted combination of means from different architectures, i.e., $\hat{\mu} \leftarrow \sum \alpha_i \mu_B^i = \sum \alpha_i f_{A_i}(x)$, where $\alpha_i$ is the sampling frequency of the $i$-th architecture.
When evaluating a specific architecture $A_i$ at test time, the estimated population statistics thus depend on the other architectures in the super-net. 
This leads to a train-test discrepancy. One solution to mitigate this problem is to re-calibrate the batch statistics by recomputing the statistics on the entire training set before the the final evaluation~\cite{yu2019slimmable}. While the cost of doing so is negligible for a standalone network, NAS algorithms typically sample $\sim 10^5$ architectures for evaluation, which makes this approach intractable.

\ky{In contrast to \cite{dong2020bench102} and \cite{Bender2020tunas} \ms{that} use the training mode also during testing,} we formalize a simple, yet effective, approach to tackle the train-test discrepancy of BN in super-net training: we leave the normalization based on batch statistics during training unchanged, but use batch statistics also during testing. Since super-net evaluation is always conducted over a complete dataset, we are free to perform inference in mini-batches of the same size as the ones used during training. This allows us to compute the batch statistics on the fly in the exact same way as during training.

\Figref{fig:bn} compares standard BN to our proposed modification. Using the tracked population statistics leads to many architectures with an accuracy around 10\%, i.e., performing no better than random guessing. Our proposed modification allows us to significantly increase the fraction of high-performing architectures.
Our results also show that the choice of fixing vs. learning an affine transformation in batch normalization should match the standalone protocol \pproxy{}.

\subsubsection{Learning rate}
\label{subsec:loss-landscape}



\begin{figure}
    \centering
    \resizebox{\linewidth}{!}{
    \includegraphics{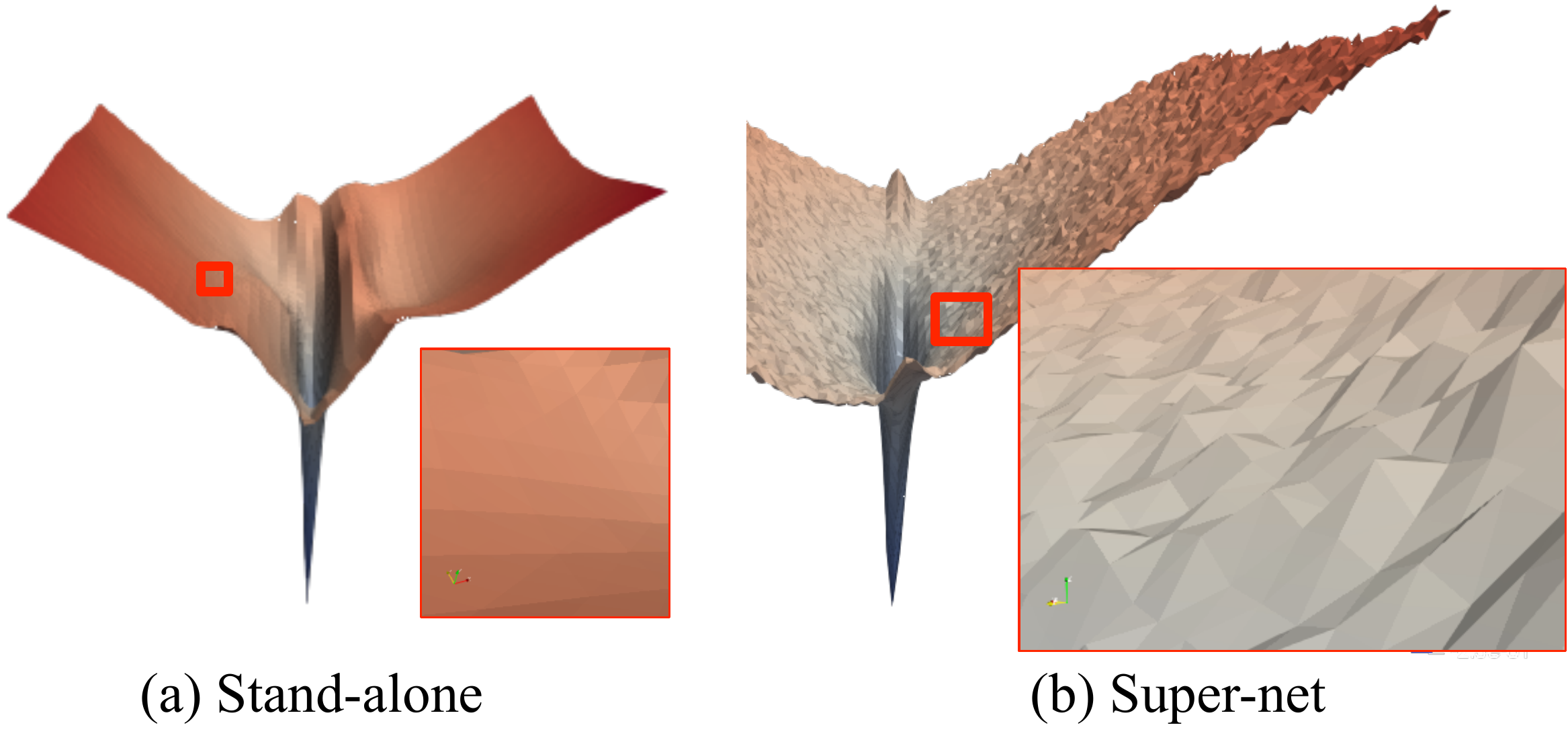}
    }
    \vspace{-0.7cm}
    \caption{\textbf{Loss landscapes.} 
    (Better see in color) of a standalone network vs the super-net ($n=300$). 
    }
    \label{fig:lr-landscape}
\end{figure}

\begin{figure}
    \centering
    \resizebox{\linewidth}{!}{
    \includegraphics{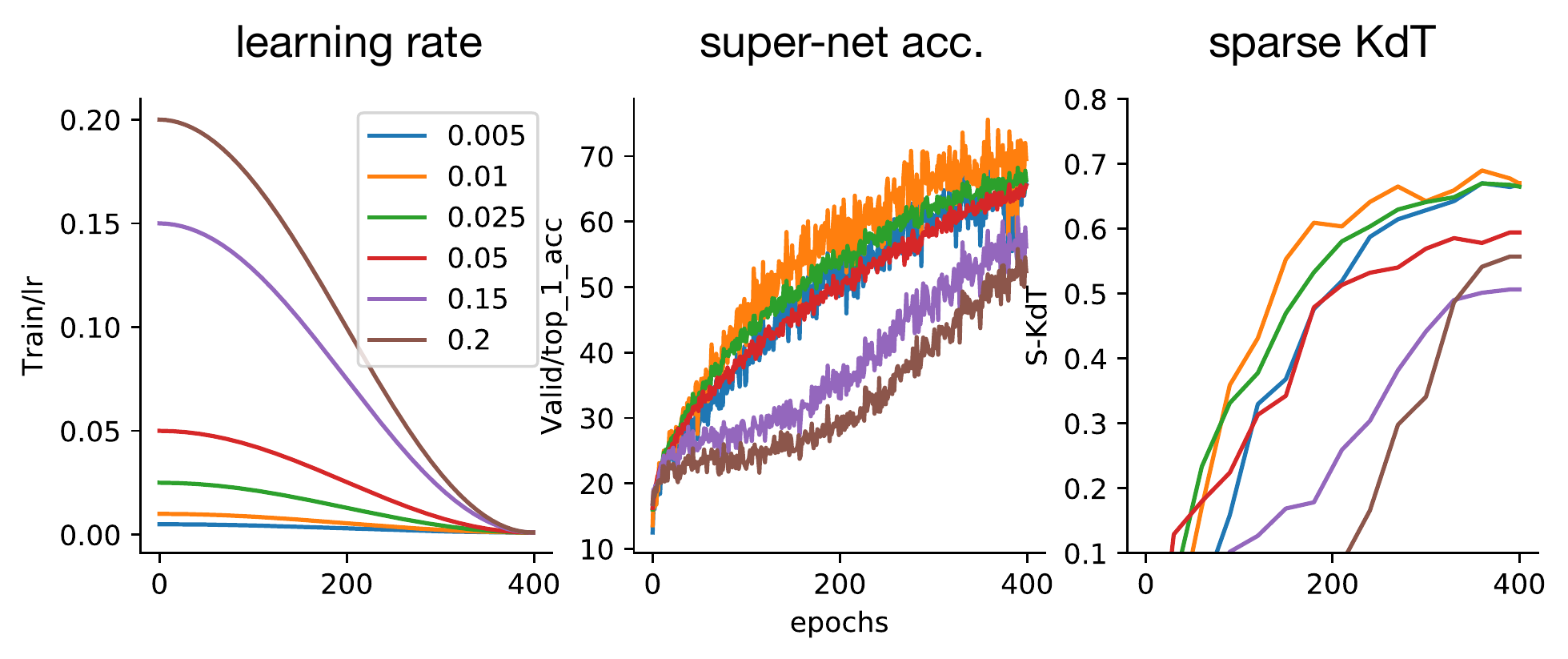}
    }
    \vspace{-0.7cm}
    \caption{Learning rate on NASBench-201. 
    }
    \label{fig:lr-nasbench201}
\end{figure}


The training loss of the super-net encompasses the task losses of all possible architectures.
We suspect that the training difficulty increases with the number of architectures represented by the super-net. To better study this, we visualize the loss landscape~\cite{li2018visualloss} of the standalone network and a super-net with $n=300$ architectures. \ky{Concretely, the landscape is computed over the super-net training loss under the single-path one-shot sampling method, \ms{i.e.,}}
\begin{equation}
\mathcal{L}_s(x, \theta_s) = \sum_i \mathcal{L}_s(x, \theta_i),\quad\text{where } \forall i, \cup_i \theta_i = \theta_s.
\end{equation}
\Figref{fig:lr-landscape} shows that the loss landscape of the super-net is less smooth than that of a standalone architecture, which confirms our intuition. \ky{A smoother landscape indicates that optimization will converge more easily to a good local optimum. With a smooth landscape, one can thus use a relatively large learning \ms{rate}. By contrast, a less smooth landscape requires using a smaller one.}

Our experiments further confirm this observation.
In the standalone protocol $P_{proxy}$, the learning rate is set to $0.2$ for NASBench-101, and to $0.1$ for NASBench-201 and DARTS-NDS, respectively. All protocols use a cosine learning rate decay.  \Figref{fig:lr-nasbench201} shows that super-net training requires lower learning rates than standalone training. The same trend is shown for other search spaces in Section~\ref{apdx:allfactor} \Tabref{tab:supp-huge-table-pws}.
\ky{We set the learning rate to 0.025 to be consistent across the three search spaces.}


\begin{figure}
    \resizebox{\linewidth}{!}{
    \includegraphics{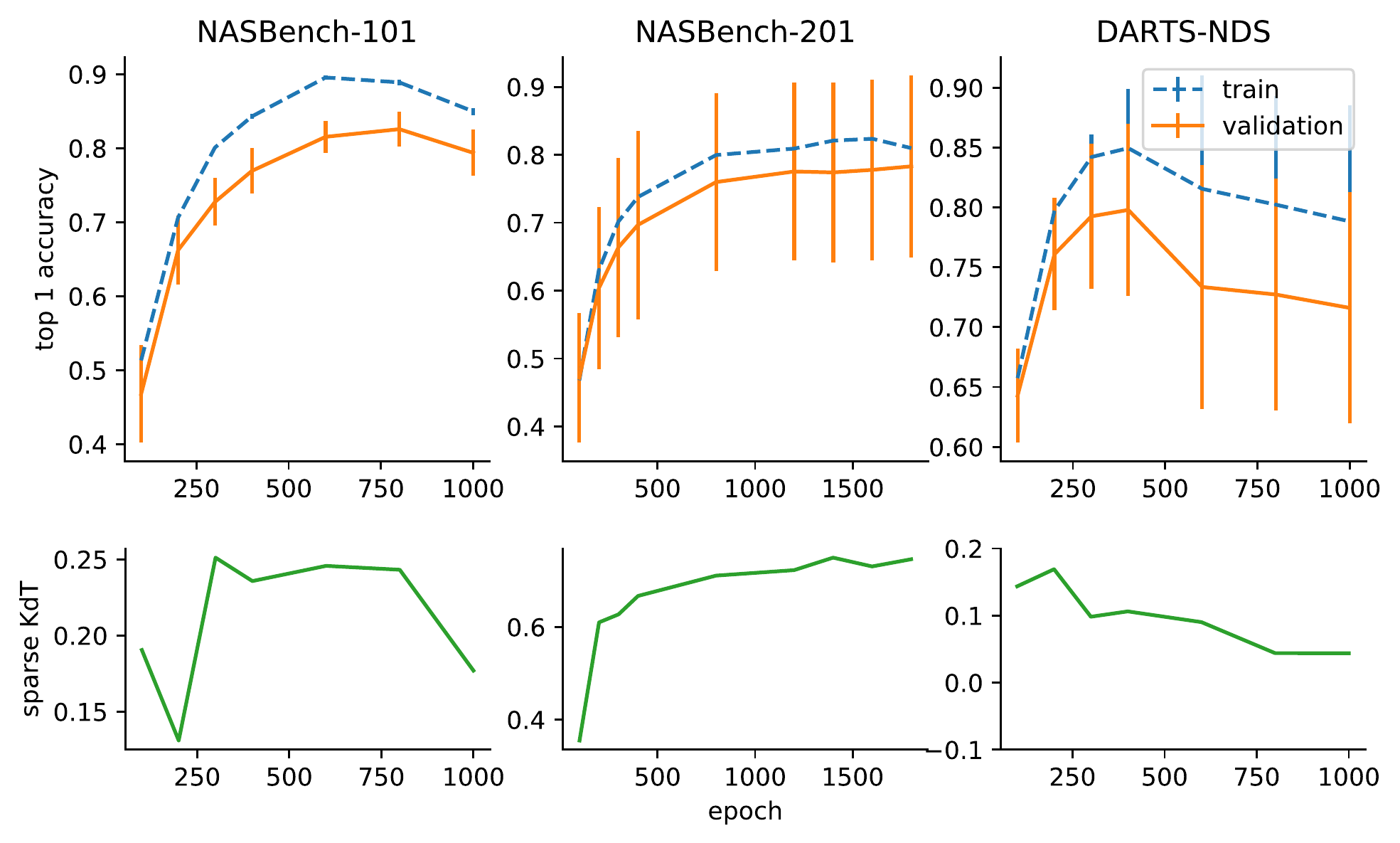}
    }
    \vspace{-0.7cm}
    \caption{\textbf{Validating the number of epochs.} 
    Each data point summarizes 3 individual runs. 
    }
    \label{fig:epochs}
    \vspace{-0.3cm}
\end{figure}
\subsubsection{Number of epochs}
Since the cosine learning rate schedule decays the learning rate to zero towards the end of  training, we evaluate the impact of the number of training epochs. In stand-alone training, the number of epochs was set to 108 for NASBench-101, 200 for NASBench-201, and 100 for DARTS-NDS. \Figref{fig:epochs} 
shows that increasing the number of epochs significantly improves the accuracy in the beginning, but eventually decreases the accuracy for NASBench-101 and DARTS-NDS. Interestingly, the number of epochs impacts neither the correlation of the ranking nor the final selected model performance after 400 epochs. 
We thus use 400 epochs for the remaining experiments.

\begin{figure}
    \centering
    \resizebox{\linewidth}{!}{
    \includegraphics{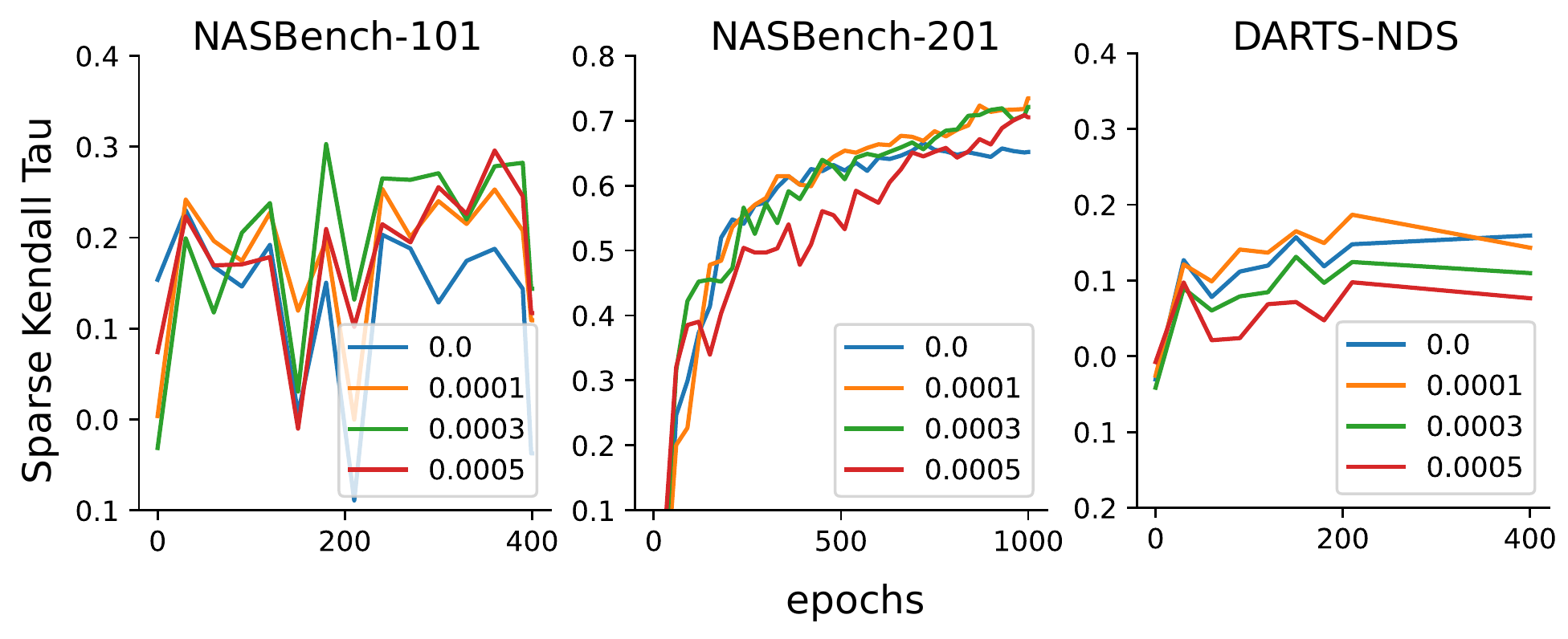}
    }
    \vspace{-0.6cm}
    \caption{\textbf{Weight decay validation.}
    }
    \vspace{-0.3cm}
    \label{fig:weight-decay}
\end{figure} 
\subsubsection{Weight decay}
Weight decay is used to reduce overfitting.
For WS-NAS, however, overfitting
does not occur because there are billions of architectures sharing the same set of parameters, which in fact rather causes underfitting.
Based on this observation,~\cite{nayman2019xnas} propose to disable weight decay during super-net training. \Figref{fig:weight-decay}, however, shows that the behavior of weight decay varies across datasets. While on DARTS-NDS weight decay is indeed harmful, it improves the results on NASBench 101 and 201. 
We conjecture that this is due to the much larger number of architectures in DARTS-NDS (243 billion) than in the NASBench series (less than 500,000).

\subsection{Weight-sharing Protocol \pws{} -- Sampling}
\label{apdx:path-sampling}

Aside from the Random-NAS described in Section~\ref{sec:method-factors}, we additionally include two variants of Random-NAS: 1) As pointed out by~\cite{ying2019bench}, two super-net architectures might be topologically equivalent in the stand-alone network by simply swapping operations. We thus include architecture-aware random sampling that ensures equal probability for unique architectures~\cite{yu2020evalnas}. We name this variant Random-A; 
2) We evaluate a variant called FairNAS \cite{chu_fairnas:_2019}, which ensures that each operation is selected with equal probability during super-net training. Although FairNAS was designed for a search space where only operations are searched but not the topology, we adapt it to our setting.

\begin{figure}
    \centering
    \resizebox{\linewidth}{!}{
    \includegraphics{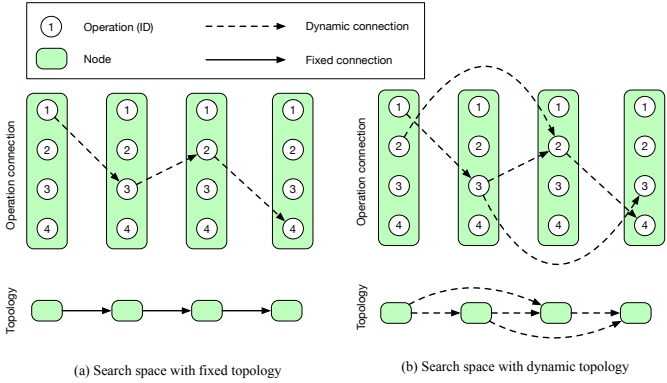}
    }
    \vspace{-0.5cm}
    \caption{Comparison between fixed and dynamic topology search spaces. }
    \label{fig:fairnas}
    \vspace{-0.3cm}
\end{figure}

\mypara{Adaptation of FairNAS.}
Originally, FairNAS~\cite{chu_fairnas:_2019} was proposed in a search space with a fixed sequential topology, as depicted by \Figref{fig:fairnas}~\captiona, where every node is sequentially connected to the previous one, and only the operations on the edges are subject to change. However, our benchmark search spaces exploit a more complex dynamic topology, as illustrated in \Figref{fig:fairnas}~\captionb, where one node can connect to one or more previous nodes. 
Before generalizing to a dynamic topology search space, we simplify the original approach into a 2-node scenario: for each input batch, FairNAS will first randomly generate a sequence of all $o$ possible operations.
It then samples one operation at a time, computes gradients for the fixed input batch, and accumulates the gradients across the operations. Once all operations have been sampled, the super-net parameters are updated with the average gradients. 
This ensures that all possible paths are equally exploited . With this simplification, FairNAS can be applied regardless of the topology. For a sequential-topology search space, we repeat the 2-node policy for every consecutive 
node pair. Naturally, for a dynamic topology space, FairNAS can be adopted in a similar manner, i.e., one first samples a topology, then applies the 2-node strategy for all connected node pairs. Note that adapting FairNAS increases the training time by a factor $o$.

\begin{figure*}[t]
    \centering
    \resizebox{\linewidth}{!}{
    \includegraphics{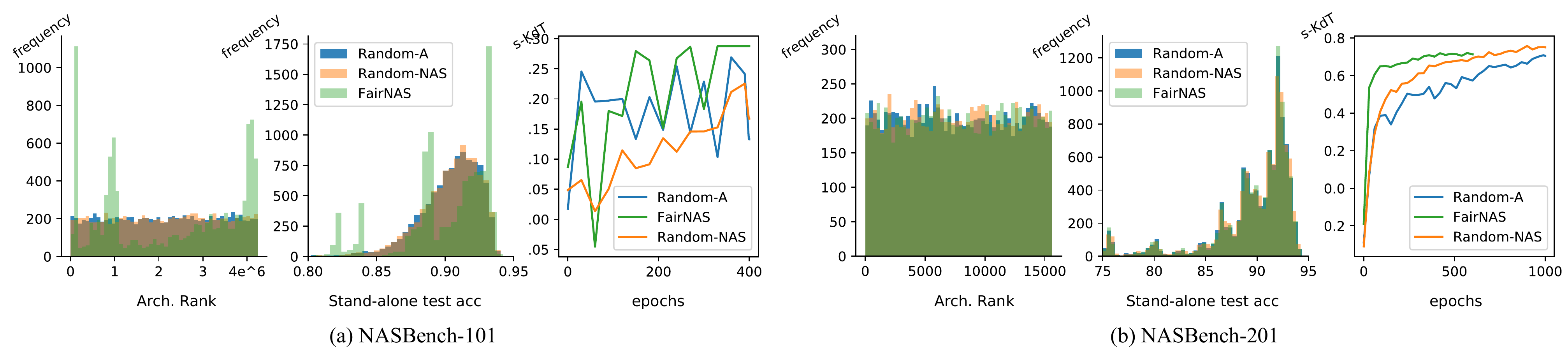}
    }
    \vspace{-0.5cm}
    \caption{
    \textbf{Path sampling comparison on NASBench-101 (a) and NASBench-201 (b).}
    We sampled 10,000 architectures using different samplers and plot histograms of the architecture rank and the stand-alone test accuracy. 
    We plot the s-KdT across the epochs.
    Results averaged across 3 runs.  
    }
    \vspace{-0.2cm}
    \label{fig:path-sample}
\end{figure*}

\mypara{Results.} With the hyper-parameters fixed, we now compare three path-sampling techniques. Since DARTS-NDS does not contain enough samples trained in a stand-alone manner, we only report results on NASBench-101 and 201. In \Figref{fig:path-sample}, we show the sampling distributions of different approaches and the impact on the super-net in terms of \skdt{}. These experiments reveal that, on NASBench-101, uniformly randomly sampling one architecture, as in~\cite{li2019random,yu2020evalnas}, is strongly biased in terms of accuracy and ranking. This can be observed from the peaks around rank 0, 100,000, and 400,000. The reason is that a single architecture can have multiple  encodings, and uniform sampling thus oversamples such architectures with
equivalent encodings.
FairNAS samples architectures more evenly and yields consistently better \skdt{}
values, albeit by a small margin.

On NASBench-201, the three sampling policies have a similar coverage. This is because, in NASBench-201, topologically-equivalent encodings were not pruned.
In this case, Random-NAS performs better than in NASBench-101, and FairNAS yields good early performance but quickly saturates. In short, using different sampling strategies might in general be beneficial, 
but we advocate for FairNAS in the presence of a limited training budget.

\subsection{Weight-sharing Mapping \fws{} -- Lower Fidelity Estimates lower the ranking correlation}

\begin{table}
    \centering
    \caption{Low fidelity estimates under same computational budget, reporting stand-alone model accuracy~(SAA) and \skdt{}~(S-KdT) on NASBench-201. }
    \vspace{-0.1cm}
    \resizebox{0.9\linewidth}{!}{
    \begin{tabular}{l|ccc}
    \toprule
    Metrics & \multicolumn{3}{c}{Settings} \\
    \midrule
    \midrule
    Batch-size &   256 & 128 & 64 \\
    \midrule 
    S-KdT & 0.740 $\pm$ 0.07 & 0.728 $\pm$ 0.16 & 0.703 $\pm$ 0.16 \\
    SAA   & 92.92 $\pm$ 0.48 & 92.37 $\pm$ 0.61 & 92.35 $\pm$ 0.34\\
    \midrule
    \midrule
     Init Channel & 16 & 8 & 4 \\
     \midrule 
     S-KdT & 0.740 $\pm$ 0.07 & 0.677 $\pm$ 0.10 & 0.691 $\pm$ 0.15 \\
     SAA   & 92.92 $\pm$ 0.48 & 92.32 $\pm$ 0.37 & 92.79 $\pm$ 0.85 \\
     \midrule
     \midrule
     Repeated cells &  3 & 2 & 1 \\
     \midrule 
     S-KdT & 0.751 $\pm$ 0.09 & 0.692 $\pm$ 0.18 & 0.502 $\pm$ 0.21\\
     SAA   & 91.91 $\pm$ 0.09 & 91.95 $\pm$ 0.10 & 90.30 $\pm$ 0.71 \\
     \midrule 
     \midrule 
     Train portion &  0.75 & 0.5 & 0.25 \\
     \midrule 
     S-KdT & 0.751 $\pm$ 0.11 & 0.742 $\pm$ 0.12 & 0.693 $\pm$ 0.13 \\
     SAA   & 92.13 $\pm$ 0.51 & 92.74 $\pm$ 0.43 & 91.47 $\pm$ 0.81 \\
     \bottomrule
    \end{tabular}
    }
    \label{tab:low-fidelity}
\end{table}
Reducing memory foot-print and training time by proposing smaller super-nets has been an active research direction, and the resulting super-nets are referred to as \textit{lower fidelity estimates}~\cite{elsken2019neural}. The impact of this approach on the super-net quality, however, has never been studied \ky{systematically over multiple search spaces}. We compare four popular strategies in 
\Tabref{tab:low-fidelity}. \ky{We deliberately prolong the training epochs inversely proportionally to the computational budget that would be saved by the low-fidelity estimates. \ms{For example, if the number of channels is reduced by half, we train the model for two times more epochs.} Note that this provides an upper bound to the performance of low-fidelity estimates.} 

A commonly-used approach to reduce memory requirements is to decrease the batch size~\cite{Yang2020NAS}. Surprisingly, lowering the batch size from 256 to 64 has limited impact on the accuracy, but decreases \skdt{} and the final searched model's performance, the most important metric in practice. 

\comment{
Another approach is to decrease the number of channels in the first layer~\cite{Liu2018darts}. 
This reduces the total number of parameters proportionally, since the number of channels in consecutive layers
depends on the first one. \Tabref{tab:low-fidelity} shows that this decreases the \skdt{} from 0.7 to 0.5. 
By contrast, reducing the number of repeated cells~\cite{Pham2018,chu_fairnas:_2019} by one has little impact. 
}

\changed{
As an alternative, one can decrease the number of  repeated cells~\cite{Pham2018,chu_fairnas:_2019}. 
This reduces the total number of parameters proportionally, since the number of cells in consecutive layers
depends on the first one. \Tabref{tab:low-fidelity} shows that this decreases the \skdt{} from 0.7 to 0.5. 
By contrast, reducing the number of channels in the first layer~\cite{Liu2018darts} has little impact. 
}
Hence, to train a good super-net, one should avoid changes between $f_{ws}$ and \fproxy{}, but one can reduce the batch size by a factor $>$ 0.5 and use only one repeated cell. 

The last lower-fidelity factor is the portion of training data that is  used~\cite{Liu2018darts,Xu2020PC-DARTS:}. Surprisingly, reducing the training portion only marginally decreases the \skdt{} for all three search spaces. On NASBench-201, keeping only 25\% of the CIFAR-10 dataset results in a 0.1 drop in \skdt{}. This explains why DARTS-based methods typically use only 50\% of the data to train the super-net but can still produce reasonable results.  

\subsection{Weight-sharing Mapping \fws{} - Implementation}
\subsubsection{Dynamic channeling hurts super-net quality}
\label{subsec:dynamic-channel}
\begin{table}[t]
    \captionof{table}{\textbf{Dynamic channels on NASBench-101.} \changed{
    We compare different ways  to handle dynamic channels in the super-net with our approach that disables dynamic channeling completely. Our approach yields a significant improvement in ranking correlation and final searched results.
    }}
    \resizebox{\linewidth}{!}{
        \begin{tabular} { l|cccc}
        \toprule
        Type & Accuracy & S-KdT & P $>$ R & Stand-alone Acc. \\
        \midrule
        Fixed       & 71.52 $\pm$ \phantom{0}6.94  & 0.22 & 0.546 & 91.79 $\pm$ 1.72 \\
        Shuffle     & 31.79 $\pm$ 10.90  & 0.17 & 0.391 & 90.58 $\pm$ 1.58 \\
        Interpolate & 57.53 $\pm$ 10.05 & 0.37 & 0.865 & 93.35 $\pm$ 3.27 \\
        \midrule
        Ours     & \textbf{76.95} $\pm$ \phantom{0}8.29 & \textbf{0.46} & \textbf{0.949} & \textbf{93.65} $\pm$ 0.73 \\
        \bottomrule
        \end{tabular}
    }
    \label{tab:dynamic-channel}
\end{table}

\begin{table}[t]
    \captionof{table}{\changed{ \textbf{Ablation study on disabling dynamic channels.} We study the effect of dynamic channeling during super-net training and during the test phase. Disabling the dynamic channels completely yields the best results.
    }}
    \resizebox{\linewidth}{!}{
        \begin{tabular} { cc|cccc}
        \toprule
         Train & Test & Accuracy & S-KdT & P $>$ R & Stand-alone Acc. \\
        \midrule
        Y & Y & 76.91 $\pm$ 10.05 & 0.22 & 0.865 & 89.43 $\pm$ 4.30 \\
        Y & N & 75.18 $\pm$ \phantom{0}9.28  & 0.33 &0.891 & 91.27 $\pm$ 1.18\\
        N & N & \textbf{76.95} $\pm$ \phantom{0}8.29 & \textbf{0.46} & \textbf{0.949} & \textbf{93.65} $\pm$ 0.73 \\
        \bottomrule
        \end{tabular}
    }
    \label{tab:dynamic-channel-ablation}
\end{table}

\begin{table}
    \caption{A fair comparison between the baseline dynamic channeling with randomly sampling sub-spaces and our disable dynamic channeling approach.} 
    \resizebox{\linewidth}{!}{
        \begin{tabular} { l|cccc}
        \toprule
        Edges & Accuracy & S-KdT & P $>$ R & Stand-alone Acc. \\
        \midrule
        \multicolumn{5}{l}{\textit{Baseline: random sampling sub-spaces with dynamic channeling.}} \\
1 &70.04$\pm$\phantom{0}8.15&0.173&0.797&91.19$\pm$2.01\\
2 &78.29$\pm$10.51&0.206&0.734&82.03$\pm$1.50\\
3 &79.92$\pm$\phantom{0}9.42&0.242&0.576&92.20$\pm$1.19\\
4 &79.37$\pm$\phantom{0}17.34&0.270&0.793&92.32$\pm$1.10\\
\midrule
Average       & 76.905 $\pm$ 10.05 & 0.223 & 0.865 & 89.435 $\pm$ 4.30 \\
        \midrule
        \midrule
        \multicolumn{5}{l}{\textit{Disable dynamic channels by fixing the edges to the output node.}} \\
1 &76.92$\pm$\phantom{0}7.87&0.435&0.991&93.94$\pm$0.22\\
2 &74.32$\pm$\phantom{0}8.21&0.426&0.925&93.34$\pm$0.01\\
3 &77.24$\pm$\phantom{0}9.18&0.487&0.901&93.66$\pm$0.07\\
4 &79.31$\pm$\phantom{0}7.04&0.493&0.978&93.65$\pm$0.07\\
\midrule
        Average       & 76.95 $\pm$ \phantom{0}8.29 & 0.460 & 0.949 & 93.65 $\pm$ 0.73 \\
        \bottomrule
        \end{tabular}
    }
    \label{tab:disable-dynamic-channel-results}
\end{table}


\begin{figure}
    \resizebox{\linewidth}{!}{
    \includegraphics{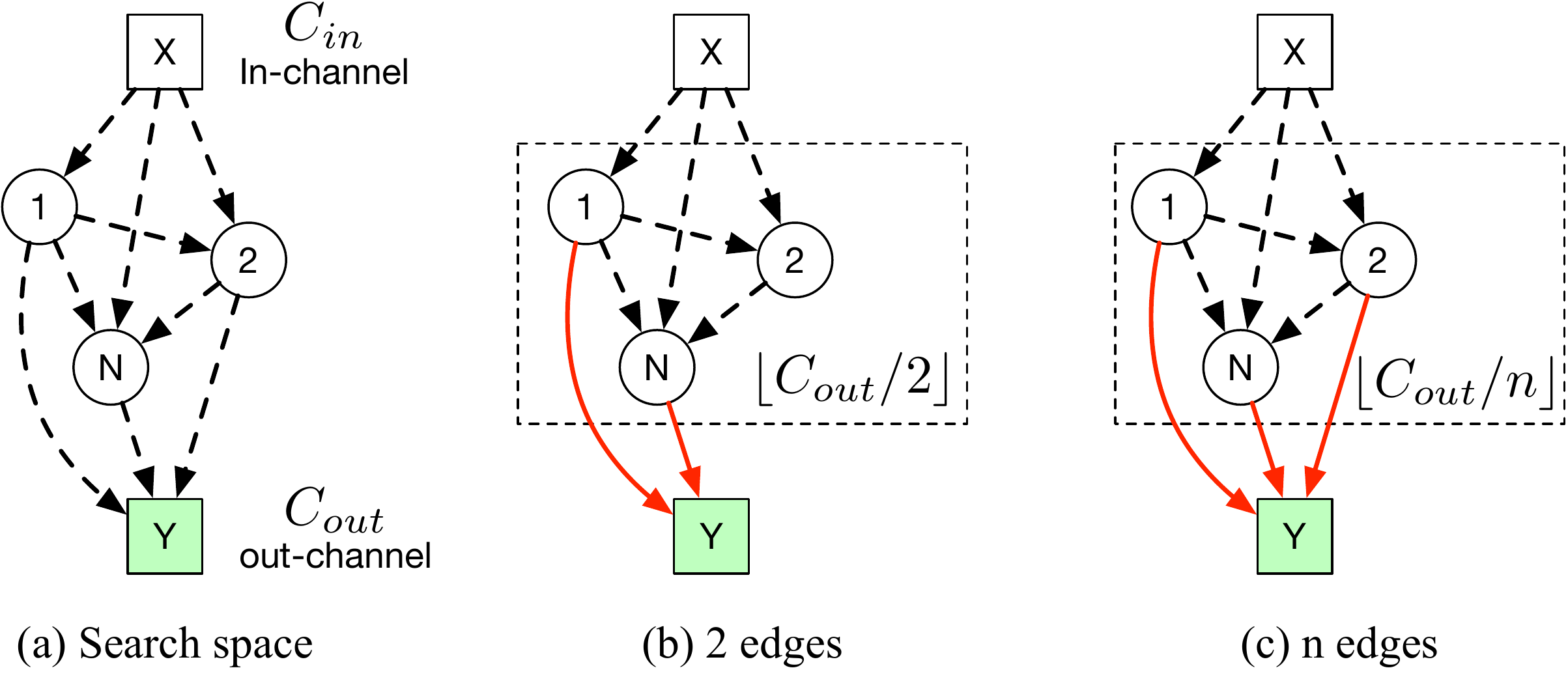}
    }
    \vspace{-0.6cm}
    \captionof{figure}{NASBench-101 dynamic channel.}
    \label{fig:disable-channel}\end{figure}
Dynamic channeling is an implicit factor in many search spaces~\cite{ying2019bench,cai2018proxyless,guo_single_2019,dong2019searching}. It refers to the fact that the number of channels of the intermediate layers depends on the number of incoming edges to the output node.
This is depicted by \Figref{fig:disable-channel}~\captiona: for a search cell with $n$ intermediate nodes, where $X$ and $Y$ are the input and output node with $C_{in}$ and $C_{out}$ channels, respectively. When there are $n=2$ edges (c.f. \Figref{fig:disable-channel}~\captionb), the associated channel numbers decrease
so that their sum equals $C_{out}$. That is, 
the intermediate nodes have $\lfloor C_{out} / 2 \rfloor$ channels. In the general case, shown in Figure \ref{fig:disable-channel}~\captionc, the number of channels in intermediate nodes is thus $\lfloor C_{out} / n \rfloor $ for $n$ incoming edges. A weight sharing approach has to cope with this architecture-dependent fluctuation of the number of channels during training.

Let $C$ denote the number of channels of a given architecture, and $C_{max}$ the maximum number of channels for a node across the entire search space. All existing approaches allocate $C_{max}$ channels and, during training, 
extract a subset of these channels. The existing methods then differ in how they extract the channels: \cite{guo_single_2019} use a fixed chunk of channels, e.g., $[0:C]$; \cite{zhang2018shufflenet} 
randomly shuffle the 
channels before extracting a fixed chunk; and \cite{dong2019network} linearly interpolate the $C_{max}$ channels into $C$ channels using a moving average across neighboring channels.


Instead of sharing the channels between architectures, we propose to disable dynamic channelling completely. As the channel number only depends on the incoming edges, we separate the search space into a discrete number of sub-spaces, each with a fixed number of incoming edges. 
As shown in \Tabref{tab:dynamic-channel}, disabling dynamic channeling improves the \skdt{} and the final search performance by a large margin and yields a new state of the art 
on NASBench101. 

\mypara{\changed{Ablation study.}}
Since each sub-space now encompasses fewer architectures, it is not fair to perform a comparison with the full NASBench 101 search space. Therefore, for each sub-space, we construct a baseline space where we drop architectures uniformly at random until the number of remaining architectures matches the size of the sub-space. We repeat this process with 3 different initializations, while keeping all other factors unchanged when training the super-net. \removed{As we enable the dynamic channels both at We refer to this as `Baseline' in \Tabref{tab:dynamic-channel-ablation}.}

We also provide additional results in \Tabref{tab:disable-dynamic-channel-results} for each individual sub-space and show that the \skdt{} remains similar to that of the baseline using the full search space, which clearly evidences the effectiveness of our approach to disable the dynamic channeling.

\changed{Furthermore, we evaluate the effect of disabling dynamic channels during the super-net training and test phase individually in \Tabref{tab:dynamic-channel-ablation}. Disabling dynamic channeling during both phases yields the best results.
}


\begin{figure*}[!ht]
    \centering
    \resizebox{\linewidth}{!}{
    \hspace{-1.cm}
    \includegraphics{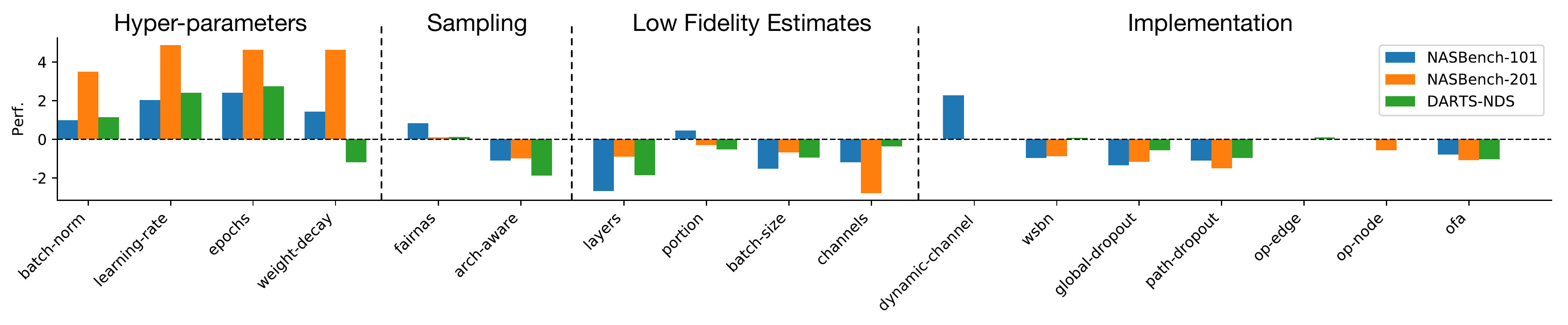}
    }
    \vspace{-0.7cm}
    \caption{\textbf{Influence of factors on the final model.} We plot the difference in percent between the searched model's performance with and without applying the corresponding factor. For the hyper-parameters of $P_{ws}$, the baseline is Random NAS, as reported in Table~\ref{tab:sota}. For the other factors, the baseline of each search space uses the best setting of the hyper-parameters. Each experiment was run at least 3 times.  
    }
    \label{fig:allfactors}
\end{figure*}

\subsubsection{WS on Edges or Nodes?}
\label{apdx:node-edge}
\begin{table}
\caption{\textbf{Comparison of operations on the nodes or on the edges.} 
We report sKT $/$ final search performance.}
    \centering
    \resizebox{0.9\linewidth}{!}{
        \begin{tabular} { l|ccc}
\toprule
& NASBench-101 & NASBench-201 & DARTS-NDS \\
\midrule
\midrule
Baseline          & 0.236 $/$ 92.32    & 0.740 $/$ 92.92    & 0.159 $/$ 93.59 \\
\midrule
Op-Edge           & N/A & as Baseline   & 0.189 $/$ 93.97 \\
Op-Node           & as Baseline         & 0.738 $/$ 92.36   & as Baseline \\
\bottomrule
\end{tabular}
}
\vspace{0.2cm} 
\label{tab:edge-node}
\resizebox{0.7\linewidth}{!}{
\includegraphics[]{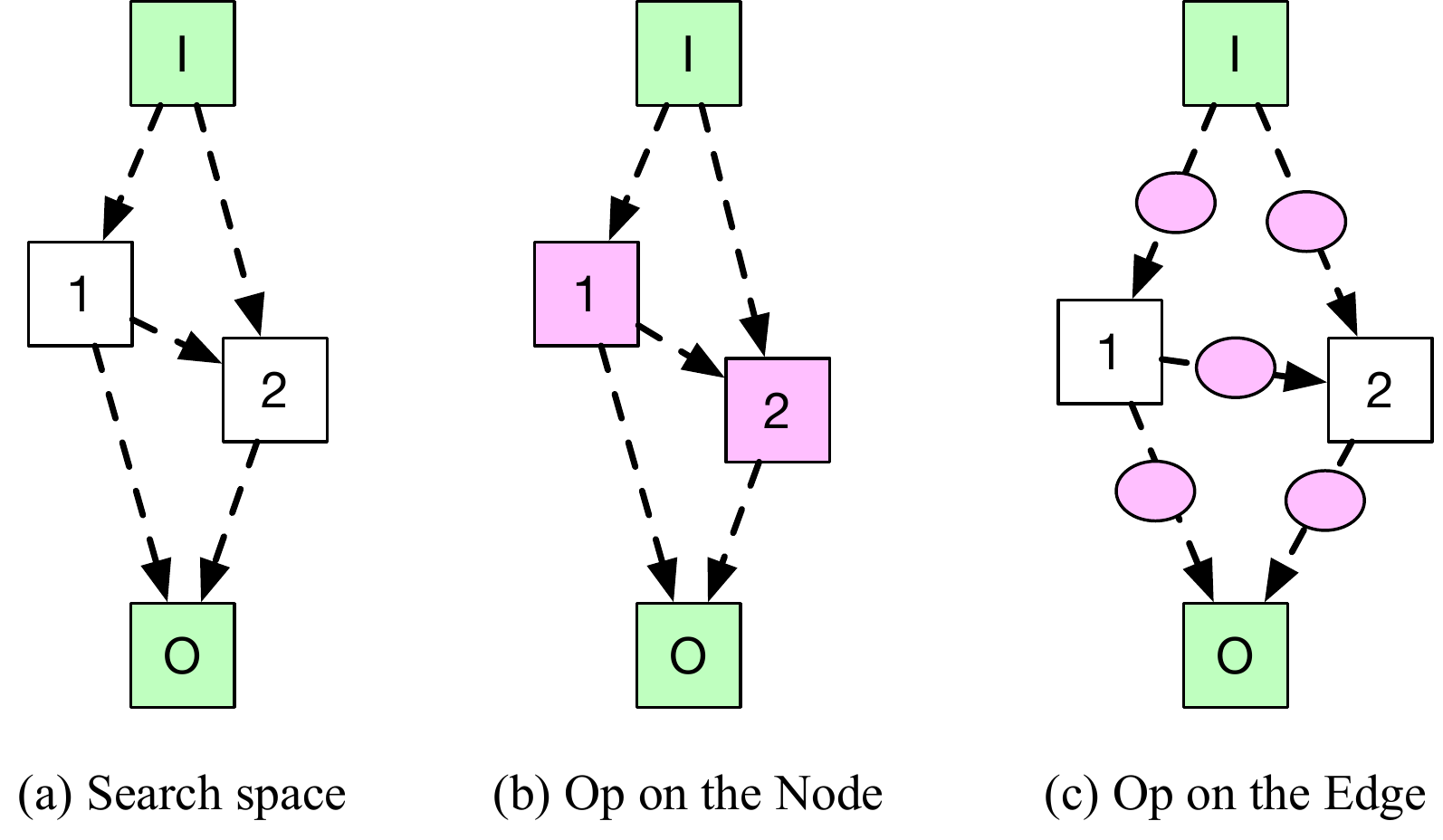}
}
\captionof{figure}{\textbf{(a)} Consider a search space with 2 intermediate nodes, 1, 2, with one input (I) and output (O) node. This yields 5 edges. Let us assume that we have 4 possible operations to choose from, as indicated as the purple color code. \textbf{(b)} When the operations are on the nodes, there are 2 $\times$ 4 ops to share, i.e., I$\rightarrow$2 and 1$\rightarrow$2 share weights on node 2. \textbf{(c)} If the operations are on the edges, then we have  5 $\times$ 4 ops to share.}
\label{fig:edge-node}
\end{table}
Most existing works build $f_{ws}$ to define the shared operations on the graph nodes rather than on the edges. This is because, if $f_{ws}$ maps to the edges, the parameter size increases from $O(n)$ to $O(n^2)$, where $n$ is the number of intermediate nodes. 
We provide a concrete example in \Figref{fig:edge-node}.
However, the high \skdt{} on NASBench-201 in the top part of \Tabref{tab:edge-node}, which is obtained by mapping to the edges, may suggest that sharing on the edges is beneficial. Here we investigate if this is truly the case.

On NASBench-101, by design, each node merges the previous nodes' outputs and then applies parametric operations. This makes it impossible to build an equivalent sharing on the edges. We therefore construct sharing on the edges for DARTS-NDS and sharing on the nodes for NASBench-201. As shown in \Tabref{tab:hp-fws}, for both spaces, sharing on the edges yields a marginally better super-net than sharing on the nodes. Such small differences might be due to the fact that, in both spaces, the number of nodes is 4, while the number of edges is 6, thus mapping to edges will not drastically affect the number of parameters. 
Nevertheless, this indicates that one should consider having a larger number of shared weights when the resources are not a bottleneck.

\subsubsection{Other mapping factors}

\begin{table}[t]
\caption{\textbf{Comparison of different mappings $f_{ws}$}. 
We report s-KdT $/$ final search performance.}
    \centering
    \resizebox{0.9\linewidth}{!}{
        \begin{tabular} { l|ccc}
\toprule
& NASBench-101 & NASBench-201 & DARTS-NDS \\
\midrule
\midrule
Baseline          & 0.236 $/$ 92.32    & 0.740 $/$ 92.92    & 0.159 $/$ 93.59 \\
\midrule
WSBN              & 0.056 $/$ 91.33    & 0.675 $/$ 92.04  & 0.331 $/$ 92.95 \\
Global-Dropout    & 0.179 $/$ 90.95     & 0.676 $/$ 91.76   & 0.102 $/$ 92.30 \\
Path-Dropout      & 0.128 $/$ 91.19     & 0.431 $/$ 91.42   & 0.090 $/$ 91.90 \\
\midrule
OFA Kernel        & 0.132 $/$ 92.01    & 0.574 $/$ 91.83  & 0.112 $/$ 92.83 \\
\bottomrule
\end{tabular}
}
\label{tab:hp-fws}
\end{table}
\begin{table}[t]
    \centering
    \caption{
        \textbf{Final results.} 
        Results on NASBench-101 and 201 are from \cite{yu2020evalnas}, and \cite{dong2020bench102}. We report the mean over 3 runs. Note that NASBench-101 ($n=7$) in \cite{yu2020evalnas} is identical to our setting. Our new strategy significantly surpasses the random search baseline.}
    \resizebox{\linewidth}{!}{
    \begin{tabular}{l|cccc}
    \toprule
             Method &  NASBench & NASBench & DARTS & DARTS \\
             & 101 (n=7) & 201 & NDS & NDS$^\star$ \\
             \midrule
             ENAS~\cite{Pham2018} & 91.83 $\pm$ 0.42 & 54.30 $\pm$ 0.00  & 94.45 $\pm$ 0.09 & 97.11 \\
             DARTS-V2~\cite{Liu2018darts} & 92.21 $\pm$ 0.61 & 54.30 $\pm$ 0.00 & 94.79 $\pm$ 0.11  & 97.37 \\
             NAO~\cite{Luo2018}  & 92.59 $\pm$ 0.59 & - & - & 97.10 \\
             GDAS~\cite{dong2019searching} & - & 93.51 $\pm$ 0.13 & - & 96.23 \\
             \midrule 
             Random NAS~\cite{li2019random} & 89.89 $\pm$ 3.89 & 87.66 $\pm$ 1.69 & 91.33 $\pm$ 0.12 & 96.74$^\dag$ \\
             Random NAS (Ours) & 93.12 $\pm$ 0.06 & 92.71 $\pm$ 0.15 & 94.26 $\pm$ 0.05 & 97.08 \\
             \bottomrule
             \multicolumn{5}{l}{$^{\dag}$Results from \cite{li2019random}} \\
             \multicolumn{5}{l}{$^{\star}$Trained according to \cite{Liu2018darts} for 600 epochs.} \\
             \multicolumn{5}{l}{\ky{On NASBench-201, both random NAS and our approach \ms{sample} 100 final}} \\
             \multicolumn{5}{l}{\ky{ architectures to follow~\cite{dong2020bench102}\ms{.}}}
    \end{tabular}
    }
    \label{tab:sota}
\end{table}
We evaluate the weight-sharing batch normalization~(WSBN) of \cite{Luo2018WSBNcode}
, which keeps an independent set of parameters for each incoming edge.
Furthermore, we test the two commonly-used dropout strategies: right before global pooling~(global dropout); and at all edge connections between the nodes~(path dropout). Note that path dropout has been widely used in WS-NAS~\cite{Luo2018,Liu2018darts,Pham2018}. For both dropout strategies, we set the dropout rate to 0.2. 
Finally, we evaluate the super convolution layer of~\cite{Cai2020Once}, referred to as OFA kernel, which accounts for the fact that, in CNN search spaces, convolution operations appear as groups, and thus merges the convolutions within the same group, keeping only the largest kernel parameters and performing a parametric projection to obtain the other kernels.
The results in ~\Tabref{tab:hp-fws} 
show that all these factors negatively impact the search performances and the super-net quality.
 
\begin{table*}[!ht]
    \centering
    \caption{\textbf{Results for all WS Protocol $P_{ws}$ factors on the three search spaces.}} 
    \vspace{-0.3cm}
    \resizebox{\textwidth}{!}{
        \begin{tabular} { l|cccc|cccc|cccc}
        \midrule[2pt]
        Factor & \multicolumn{4}{c}{NASBench-101} & \multicolumn{4}{c}{NASBench-201} & \multicolumn{4}{c}{DARTS-NDS} \\
        and & Super-net & \phantom{0} & \phantom{0} & Final & Super-net & \phantom{0} & \phantom{0} & Final & Super-net & \phantom{0} &  \phantom{0} & Final \\
        settings & Accuracy & S-KdT & P $>$ R &  Performance & Accuracy & S-KdT & P $>$ R & Performance & Accuracy & S-KdT & P $>$ R & Performance \\
        \midrule[1pt]
        \multicolumn{3}{l}{\textbf{Batch-norm. }} \\
        \midrule
affine F track F &0.651$\pm$0.05&0.161&0.996&0.916$\pm$0.13&0.660$\pm$0.13&0.783&0.997&92.67$\pm$1.21&0.735$\pm$0.18&0.056&0.224&93.14$\pm$0.28\\
affine T track F &0.710$\pm$0.04&0.240&0.996&0.924$\pm$0.01&0.713$\pm$0.14&0.718&0.707&91.71$\pm$1.05&0.265$\pm$0.21&-0.071&0.213&91.89$\pm$2.01\\
affine F track T &0.144$\pm$0.09&0.084&0.112&0.882$\pm$0.02&0.182$\pm$0.15&-0.171&0.583&86.41$\pm$4.84&0.359$\pm$0.25&-0.078&0.023&90.33$\pm$0.76\\
affine T track T &0.153$\pm$0.10&-0.008&0.229&0.905$\pm$0.01&0.134$\pm$0.09&-0.417&0.274&90.77$\pm$0.40&0.216$\pm$0.18&-0.050&0.109&90.49$\pm$0.32\\
        \midrule[1pt]
        \multicolumn{3}{l}{\textbf{Learning rate.}} \\
        \midrule
0.005 &0.627$\pm$0.07&0.091&0.326&0.908$\pm$0.01&0.658$\pm$0.11&0.668&0.141&90.14$\pm$0.55 & 0.792$\pm$0.08&0.130&0.033&91.81$\pm$0.68\\
0.01 &0.668$\pm$0.06&0.095&0.546&0.919$\pm$0.00&0.713$\pm$0.12&0.670&0.711&91.21$\pm$1.18 & 0.727$\pm$0.05&0.131&0.258&92.86$\pm$0.64\\
0.025 &0.715$\pm$0.05&0.220&0.910&0.917$\pm$0.01&0.659$\pm$0.13&0.665&0.844&92.42$\pm$0.58 & 0.656$\pm$0.14&0.218&0.299&93.42$\pm$0.20\\
0.05 &0.727$\pm$0.05&0.143&0.905&0.911$\pm$0.02&0.631$\pm$0.14&0.594&0.730&92.02$\pm$0.70&0.623$\pm$0.04&0.147&0.489&91.70$\pm$0.33\\
0.1 &0.690$\pm$0.07&0.005&0.905&0.909$\pm$0.02&0.609$\pm$0.28&0.571&0.618&91.82$\pm$0.81&0.735$\pm$0.06&0.096&0.099&92.73$\pm$0.24\\
0.15 &0.000$\pm$0.00&-0.274 & N/A & N/A &0.551$\pm$0.14&0.506&0.553&91.22$\pm$1.20&0.371$\pm$0.27&0.027&0.218&91.20$\pm$0.72\\
0.2 &-&-&-&-&0.519$\pm$0.12&0.557&0.035&88.74$\pm$0.11&0.102$\pm$0.48&-0.366&N/A&N/A\\
        \midrule[1pt]
        \multicolumn{10}{l}{\textbf{Epochs.}} \\
        \midrule
100 &0.468$\pm$0.07&0.190&0.759&0.920$\pm$0.01&0.472$\pm$0.09&0.355&0.997&92.11$\pm$1.67&0.643$\pm$0.04&0.144&0.901&93.90$\pm$0.49\\
200 &0.662$\pm$0.05&0.131&0.685&0.914$\pm$0.01&0.604$\pm$0.12&0.610&0.881&91.88$\pm$2.01&0.761$\pm$0.05&0.169&0.778&94.08$\pm$0.21\\
300 &0.727$\pm$0.03&0.251&0.739&0.920$\pm$0.01&0.664$\pm$0.13&0.627&0.840&91.42$\pm$1.91&0.793$\pm$0.06&0.098&0.870&93.22$\pm$0.95\\
400 &0.769$\pm$0.03&0.236&0.932&0.921$\pm$0.01&0.697$\pm$0.14&0.667&0.158&89.83$\pm$0.97&0.798$\pm$0.07&0.106&0.036&92.34$\pm$0.22\\
600 &0.815$\pm$0.02&0.246&0.556&0.911$\pm$0.01&0.720$\pm$0.13&0.682&0.285&90.28$\pm$0.82&0.734$\pm$0.10&0.090&0.209&93.23$\pm$0.19\\
800 &0.826$\pm$0.02&0.243&0.177&0.907$\pm$0.00&0.760$\pm$0.13&0.711&0.378&91.53$\pm$0.53&0.728$\pm$0.10&0.044&0.853&93.29$\pm$0.81\\
1000 &0.794$\pm$0.03&0.177&0.831&0.920$\pm$0.01&0.782$\pm$0.13&0.740&0.589&92.92$\pm$0.48&0.717$\pm$0.09&0.044&0.997&93.92$\pm$0.90\\
1200 &-&-&-&-&0.775$\pm$0.13&0.723&0.198&90.81$\pm$0.56&-&-&-&-\\
1400 &-&-&-&-&0.774$\pm$0.13&0.750&0.604&92.26$\pm$0.33&-&-&-&-\\
1600 &-&-&-&-&0.778$\pm$0.13&0.731&0.882&91.85$\pm$1.20&-&-&-&-\\
1800 &-&-&-&-&0.783$\pm$0.13&0.746&0.266&90.64$\pm$0.82&-&-&-&-\\
        \midrule[1pt]
        \multicolumn{10}{l}{\textbf{Weight decay.} } \\
        \midrule
0.0 &0.645$\pm$0.05&-0.037&0.179&0.899$\pm$0.01&0.713$\pm$0.13&0.652&0.266&90.58$\pm$0.99&0.670$\pm$0.03&0.159&0.629&93.09$\pm$0.73\\
0.0001 &0.719$\pm$0.03&0.109&0.659&0.912$\pm$0.01&0.756$\pm$0.13&0.734&0.612&91.88$\pm$0.59&0.751$\pm$0.05&0.143&0.396&93.37$\pm$0.44\\
0.0003 &0.771$\pm$0.03&0.144&0.648&0.915$\pm$0.01&0.772$\pm$0.13&0.721&0.726&92.34$\pm$0.57&0.759$\pm$0.06&0.110&0.890&93.82$\pm$0.51\\
0.0005 &0.782$\pm$0.03&0.117&0.910&0.911$\pm$0.02&0.764$\pm$0.13&0.705&0.882&92.61$\pm$0.59&0.739$\pm$0.07&0.077&0.051&91.61$\pm$1.01\\
        \midrule[1pt]
\multicolumn{10}{l}{\textbf{Sampling.} } \\
\midrule
Random-A &0.717$\pm$0.04&0.133&0.862&0.919$\pm$0.02&0.764$\pm$0.13&0.705&0.882&92.61$\pm$0.59&-&-&-&-\\
Random-NAS &0.638$\pm$0.20&0.167&0.949&0.913$\pm$0.02&0.765$\pm$0.14&0.750&0.897&92.17$\pm$1.01&-&-&-&-\\
FairNAS &0.789$\pm$0.03&0.288&0.382&0.908$\pm$0.01&0.774$\pm$0.14&0.713&0.917&93.06$\pm$0.31&-&-&-&-\\
        \midrule[2pt]
        \end{tabular}
    }
\label{tab:supp-huge-table-pws}
\end{table*}

\begin{table*}[!ht]
	\caption{\textbf{Results for all low-fidelity factors on the three search spaces.}} 
	\vspace{-0.3cm}
\resizebox{\textwidth}{!}{
	\begin{tabular} { l|cccc|cccc|cccc}
        \midrule[2pt]
		Factor & \multicolumn{4}{c}{NASBench-101} & \multicolumn{4}{c}{NASBench-201} & \multicolumn{4}{c}{DARTS-NDS} \\
		and & Super-net & \phantom{0} & \phantom{0} & Final & Super-net & \phantom{0} & \phantom{0} & Final & Super-net & \phantom{0} &  \phantom{0} & Final \\
		settings & Accuracy & S-KdT & P $>$ R &  Performance & Accuracy & S-KdT & P $>$ R & Performance & Accuracy & S-KdT & P $>$ R & Performance \\
        \midrule[1pt]
		\multicolumn{10}{l}{\textbf{Number of Layer} (-X indicates the baseline minus X)} \\
		\midrule
		Baseline & 0.769$\pm$0.03&0.236&0.932&0.921$\pm$0.01 & 0.782$\pm$0.13&0.740&0.589&92.92$\pm$0.48 & 0.670$\pm$0.03&0.159&0.629&93.09$\pm$0.73 \\
		-1 &0.759$\pm$0.03&0.214&0.222&0.901$\pm$0.01 & 0.749$\pm$0.13&0.710&0.796&91.85$\pm$0.92 &0.843$\pm$0.04&0.178&0.299&92.35$\pm$1.25 \\
		-2 & 0.817$\pm$0.03&0.228&0.713&0.910$\pm$0.02 & 0.777$\pm$0.13&0.700&0.822&92.68$\pm$0.37 & 0.852$\pm$0.03&0.205&0.609&92.65$\pm$1.89\\
        \midrule[1pt]
		\multicolumn{3}{l}{\textbf{Train portion}} \\
		\midrule
		0.25 &0.433$\pm$0.07&0.216&0.281&0.901$\pm$0.01&0.660$\pm$0.11&0.668&0.979&92.30$\pm$1.14&0.597$\pm$0.14&0.132&0.359&92.27$\pm$1.84\\
		0.5 &0.612$\pm$0.06&0.251&0.424&0.896$\pm$0.02&0.740$\pm$0.12&0.669&0.979&93.17$\pm$0.47&0.666$\pm$0.17&0.083&0.551&92.22$\pm$1.36\\
		0.75 &0.688$\pm$0.05&0.222&0.857&0.920$\pm$0.01&0.758$\pm$0.13&0.725&0.618&92.46$\pm$0.19&0.715$\pm$0.18&0.096&0.081&92.29$\pm$0.47\\
		0.9 &0.722$\pm$0.05&0.186&0.996&0.931$\pm$0.01&0.772$\pm$0.13&0.721&0.726&92.34$\pm$0.57&0.703$\pm$0.18&0.042&0.065&92.78$\pm$0.10\\
        \midrule[1pt]
		\multicolumn{10}{l}{\textbf{Batch size} (/ X indicates the baseline divide by X)} \\
		\midrule
		Baseline & 0.769$\pm$0.03&0.236&0.932&0.921$\pm$0.01 & 0.782$\pm$0.13&0.740&0.589&92.92$\pm$0.48 & 0.670$\pm$0.03&0.159&0.629&93.09$\pm$0.73 \\
		/ 2 & 0.670$\pm$0.05&0.246&0.807&0.920$\pm$0.01 & 0.728$\pm$0.16&0.719&0.842&92.37$\pm$0.61 &0.698$\pm$0.20&0.037&0.209&93.24$\pm$0.13\\
		/ 4 &0.686$\pm$0.07&0.155&0.913&0.921$\pm$0.01&0.703$\pm$0.16&0.679&0.672&92.35$\pm$0.34 & 0.633$\pm$0.20&0.033&0.690&93.68$\pm$0.62\\
        \midrule[1pt]
		\multicolumn{10}{l}{\textbf{\# channel} (/ X indicates the baseline divide by X)} \\
		\midrule
		Baseline & 0.769$\pm$0.03&0.236&0.932&0.921$\pm$0.01 & 0.782$\pm$0.13&0.740&0.589&92.92$\pm$0.48 & 0.670$\pm$0.03&0.159&0.629&93.09$\pm$0.73 \\
		/ 2 &0.658$\pm$0.05&0.156&0.704&0.898$\pm$0.02 & 0.697$\pm$0.14&0.667&0.158&89.83$\pm$0.97 & 0.776$\pm$0.05&0.190&0.993&93.90$\pm$0.71\\
		/ 4 &0.604$\pm$0.06&0.093&0.907&0.922$\pm$0.01& 0.606$\pm$0.13&0.616&0.878&92.86$\pm$0.34 &0.707$\pm$0.05&0.202&0.359&92.93$\pm$0.58 \\
        \midrule[2pt]
	\end{tabular}
}
\label{tab:supp-huge-table-low-fidelity}
\end{table*}
\vspace{-0.2cm}
\begin{table*}[!ht]
	\caption{\textbf{Results for all implementation factors on the three search spaces.}} 
	\vspace{-0.3cm}
\resizebox{\textwidth}{!}{
	\begin{tabular} { l|cccc|cccc|cccc}
        \midrule[2pt]
		Factor & \multicolumn{4}{c}{NASBench-101} & \multicolumn{4}{c}{NASBench-201} & \multicolumn{4}{c}{DARTS-NDS} \\
		and & Super-net & \phantom{0} & \phantom{0} & Final & Super-net & \phantom{0} & \phantom{0} & Final & Super-net & \phantom{0} &  \phantom{0} & Final \\
		settings & Accuracy & S-KdT & P $>$ R &  Performance & Accuracy & S-KdT & P $>$ R & Performance & Accuracy & S-KdT & P $>$ R & Performance \\
        \midrule[1pt]
		\multicolumn{10}{l}{\textbf{Other factors}} \\
		\midrule
				Baseline & 0.769$\pm$0.03&0.236&0.932&0.921$\pm$0.01 & 0.782$\pm$0.13&0.740&0.589&92.92$\pm$0.48 & 0.670$\pm$0.03&0.159&0.629&93.09$\pm$0.73 \\
		OFA Kernel & 0.708$\pm$0.08 & 0.132 & 0.203 & 92.01$\pm$0.19 & 0.672$\pm$0.18 &0.574 & 0.605 & 91.83 $\pm$ 0.86& 0.782$\pm$0.05&0.112&0.399&93.22$\pm$0.43 \\
		WSBN  &0.155$\pm$0.07&0.085&0.504&0.809$\pm$0.13&0.703$\pm$0.14&0.676&0.585&92.06$\pm$0.48&0.744$\pm$0.16&0.033&0.682&92.88$\pm$1.22\\
        \midrule[1pt]
		\multicolumn{10}{l}{\textbf{Path dropout rate}} \\
		\midrule
			Baseline & 0.769$\pm$0.03&0.236&0.932&0.921$\pm$0.01 & 0.782$\pm$0.13&0.740&0.589&92.92$\pm$0.48 & 0.670$\pm$0.03&0.159&0.629&93.09$\pm$0.73 \\
		0.05 &0.750$\pm$0.02&0.206&0.819&0.915$\pm$0.07 &0.490$\pm$0.09&0.712&0.881&92.25$\pm$0.89&0.184$\pm$0.06&0.006&0.359&92.93$\pm$0.60\\
		0.15 &0.726$\pm$0.02&0.186&0.482&0.910$\pm$0.01&0.250$\pm$0.03&0.640&0.526&91.44$\pm$1.25&0.366$\pm$0.05&0.059&0.570&92.61$\pm$1.28\\
		0.2 &0.669$\pm$0.01&0.110&0.282&0.901$\pm$0.01&0.185$\pm$0.02&0.431&0.809&92.15$\pm$0.85&0.518$\pm$0.06&0.090&0.009&91.45$\pm$0.58\\
        \midrule[1pt]
		\multicolumn{10}{l}{\textbf{Global dropout}} \\
		\midrule
		Baseline & 0.769$\pm$0.03&0.236&0.932&0.921$\pm$0.01 & 0.782$\pm$0.13&0.740&0.589&92.92$\pm$0.48 & 0.670$\pm$0.03&0.159&0.629&93.09$\pm$0.73 \\
		0.2 &0.739$\pm$0.05&0.233&0.221&0.910$\pm$0.00&0.712$\pm$0.13&0.702&0.950&91.76$\pm$1.36&0.557$\pm$0.19&0.018&0.451&93.51$\pm$0.27\\
        \midrule[2pt]
       \multicolumn{10}{l}{Please refer to Section~\ref{apdx:node-edge} for mapping on the node or edge and Section~\ref{subsec:dynamic-channel} for dynamic channel factor results.}
	\end{tabular}
}
\label{tab:supp-huge-table-implementation}
\end{table*}

\subsection{Results for All Factors}
\label{apdx:allfactor}
We report the numerical results for all hyper-parameter factors in \Tabref{tab:supp-huge-table-pws}, low-fidelity factors in \Tabref{tab:supp-huge-table-low-fidelity} and implementation factors in \Tabref{tab:supp-huge-table-implementation}. These results were computed from the last epochs of 3 different runs.

\section{How should you train your super-net?}
\label{sec:discussion}

\Figref{fig:allfactors} summarizes the influence of all tested factors on the final performance. 
It stands out that properly tuned hyper-parameters lead to the biggest improvements by far. Surprisingly, most other factors and techniques either have a hardly measurable effect or in some cases even lead to worse performance. 
Based on these findings, 
here is how you should train your super-net: 
\begin{enumerate}[noitemsep,topsep=0pt,parsep=0pt,partopsep=0pt,leftmargin=12pt]
    \item Do not use super-net accuracy to judge the quality of your super-net. The \skdt{} has much higher correlation with the final search performance. 
    \item When batch normalization is used, do not use the moving average statistics during evaluation. Instead, compute the statistics on the fly over a batch of the same size as used during training.
    \item The loss landscape of super-nets is less smooth than that of standalone networks. Start from a smaller learning rate than standalone training. 
    \item Do not use other low-fidelity estimates than moderately reducing the training set size to decrease the search time.
    \item Do not use dynamic channeling in search spaces that have a varying number of channels in the intermediate nodes. Break the search space into multiple sub-spaces such that dynamic channeling is not required.
\end{enumerate}

\mypara{Comparison to the state of the art.}
\Tabref{tab:sota} shows that carefully controlling the relevant factors and adopting the techniques proposed in \Secref{sec:analysis} allow us to considerably improve the performance of Random-NAS. 
Thanks to our evaluation, we were able to show that simple Random-NAS together with an appropriate training protocol $P_{ws}$ and mapping function $f_{ws}$ yields results that are competitive to and sometimes even surpass state-of-the-art algorithms. 
Our results provide a strong baseline upon which future work can build.

We also report the best settings in ~\Tabref{tab:best-setting}.
\begin{table*}[!ht]
    \centering
    \caption{Parameter settings that obtained the best searched results. }
    \vspace{-0.2cm}
    \resizebox{\textwidth}{!}{
    \begin{tabular}{l|ccccc|cccc|cccc|c}
    \toprule
        Search Space & \multicolumn{5}{c}{\textit{implementation}} & \multicolumn{4}{c}{\textit{low fidelity}} & \multicolumn{4}{c}{\textit{hyperparam.}} & \textit{sampling} \\
        \cmidrule{1-5} \cmidrule{6-9} \cmidrule{10-13} \cmidrule{14-15}
        & Dynamic Conv & OFA Conv & WSBN & Dropout & Op map & $\#$ layer & portion & batch-size & $\#$ channels & batch-norm & learning rate & epochs & weight decay &  \\
    \midrule
    NASBench-101 & Interpolation & N  & N & 0. & Node & 9 & 0.75 & 256 & 128 & Tr=F A=T & 0.025 & 400 & 1e-3 & FairNAS  \\
    NASBench-201 & Fix & N  & N & 0. & Edge & 5 & 0.9 & 128 & 16 &  Tr=F A=T & 0.025 & 1000 & 3e-3 & FairNAS  \\
    DARTS-NDS & Fix & N  & Y & 0. & Edge & 12 & 0.9 & 256 & 36 & Tr=F A=F & 0.025 & 400 & 0 & FairNAS  \\
    \bottomrule
    \multicolumn{15}{l}{For batch-norm, we report Track statistics (Tr) and Affine (A) setting with True (T) or False (F).}\\
    \multicolumn{15}{l}{For other notation, Y = Yes, N = No.}
    \end{tabular}
    }
    \label{tab:best-setting}
\end{table*}

\ifCLASSOPTIONcompsoc
  \section*{Acknowledgments}
\else
  \section*{Acknowledgment}
\fi

This work was partially done during an internship at Intel and at Abacus.AI, and supported in part by the Swiss National Science Foundation.

\ifCLASSOPTIONcaptionsoff
  \newpage
\fi

\begin{IEEEbiography}[{\includegraphics[width=1in,height=1.25in,clip,keepaspectratio]{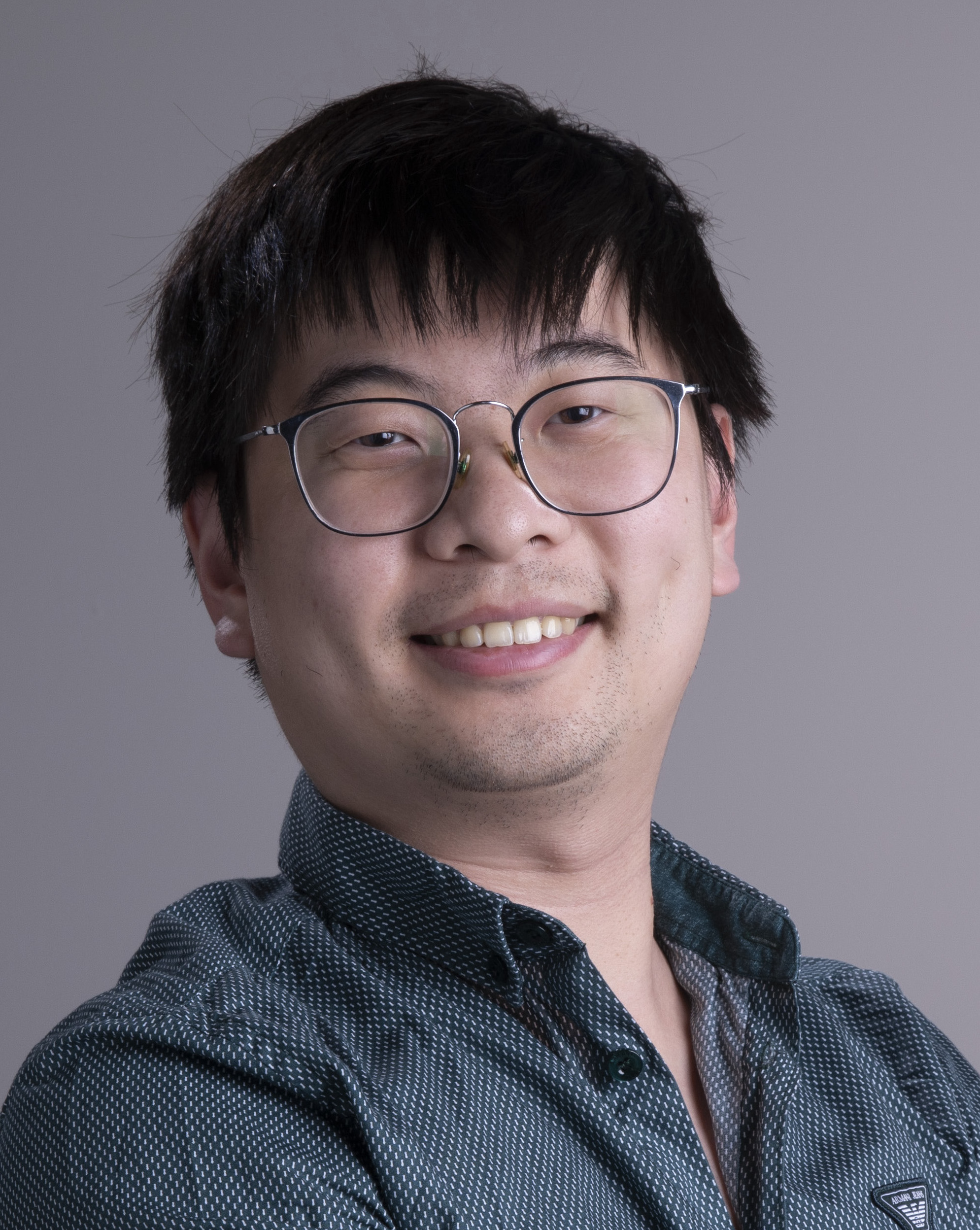}}]{Kaicheng Yu}
is a PhD candidate at EPFL and works with Abacus.AI. His research interests spans in computer vision and machine learning. Kaicheng is a recipient of Qualcomm Innovation Fellowship (Europe) 2019. He received his Bachelor in Computer Science from the University of Hong Kong in 2016.
\vspace{-12mm}
\end{IEEEbiography}

\begin{IEEEbiography}[{\includegraphics[width=1in,height=1.25in,clip,keepaspectratio]{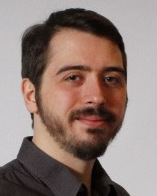}}]{Ren\'{e} Ranftl}
is a Senior Research Scientist at the Intelligent Systems Lab at Intel in Munich, Germany. He received an M.Sc. degree and a Ph.D. degree from Graz University of Technology, Austria, in 2010 and 2015, respectively. His research interests broadly span topics in computer vision, machine learning, and robotics.
\end{IEEEbiography}
\vspace{-12mm}

\begin{IEEEbiography}[{\includegraphics[width=1in,height=1.25in,clip,keepaspectratio]{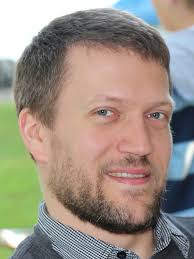}}]{Mathieu Salzmann}
is a Senior Researcher at EPFL and an Artificial Intelligence Engineer at ClearSpace. Previously, he was a Senior Researcher and Research Leader in NICTA's computer vision research group, a Research Assistant Professor at TTI-Chicago, and a postdoctoral fellow at ICSI
and EECS at UC Berkeley. He obtained his PhD in 2009 from EPFL. His research interests lie at the intersection of machine learning and computer vision. 
\end{IEEEbiography}




\end{document}